\documentclass[10pt,twocolumn,letterpaper]{article}

\usepackage{iccv}              

\definecolor{iccvblue}{rgb}{0.21,0.49,0.74}
\usepackage[pagebackref,breaklinks,colorlinks,allcolors=iccvblue]{hyperref}

\usepackage{amsmath}  
\usepackage{amsthm}  
\usepackage{graphicx}  
\usepackage{tcolorbox} 
\usepackage{multirow}
\usepackage{colortbl}
\usepackage{bbding}
\usepackage{wrapfig}
\usepackage{array}  

\usepackage{amssymb,booktabs}
\usepackage{tabularx}
\usepackage{bbding}
\usepackage{pifont}
\usepackage{makecell}
\usepackage{float}

\def\paperID{*****} 
\def\confName{ICCV}
\def\confYear{2025}

\title{RemoteSAM: Towards Segment Anything for Earth Observation}

\author{
\textbf{Liang Yao}\textsuperscript{1,*},
\textbf{Fan Liu}\textsuperscript{1,*,†},
\textbf{Delong Chen}\textsuperscript{2,*},
\textbf{Chuanyi Zhang}\textsuperscript{1}\\
\textbf{Yijun Wang}\textsuperscript{1},
\textbf{Ziyun Chen}\textsuperscript{1},
\textbf{Wei Xu}\textsuperscript{1},
\textbf{Shimin Di}\textsuperscript{3},
\textbf{Yuhui Zheng}\textsuperscript{1}
\\\\
\textsuperscript{1}Hohai University \quad
\textsuperscript{2}HKUST \quad
\textsuperscript{3}Southeast University \\
\\
{\tt\small \textsuperscript{*}Equal Contribution \quad \textsuperscript{†}Corresponding Author
}
\\
{\tt\small Email: fanliu@hhu.edu.cn}
}

\begin{document}
\maketitle
\begin{abstract}
We aim to develop a robust yet flexible visual foundation model for Earth observation. It should possess strong capabilities in recognizing and localizing diverse visual targets while providing compatibility with various input-output interfaces required across different task scenarios. Current systems cannot meet these requirements, as they typically utilize task-specific architecture trained on narrow data domains with limited semantic coverage. Our study addresses these limitations from two aspects: data and modeling. 
We first introduce an automatic data engine that enjoys significantly better scalability compared to previous human annotation or rule-based approaches. It has enabled us to create the largest dataset of its kind to date, comprising 270K image-text-mask triplets covering an unprecedented range of diverse semantic categories and attribute specifications. 
Based on this data foundation, we further propose a task unification paradigm that centers around referring expression segmentation. It effectively handles a wide range of vision-centric perception tasks, including classification, detection, segmentation, grounding, etc, using a single model without any task-specific heads. 
Combining these innovations on data and modeling, we present RemoteSAM, a foundation model that establishes new SoTA on several earth observation perception benchmarks, outperforming other foundation models such as Falcon, GeoChat, and LHRS-Bot with significantly higher efficiency. Models and data are publicly available at \url{https://github.com/1e12Leon/RemoteSAM}.  
\end{abstract}    
\section{Introduction}
\label{sec:intro}

\begin{figure}[t]
  \centering
  \includegraphics[width=0.99\linewidth]{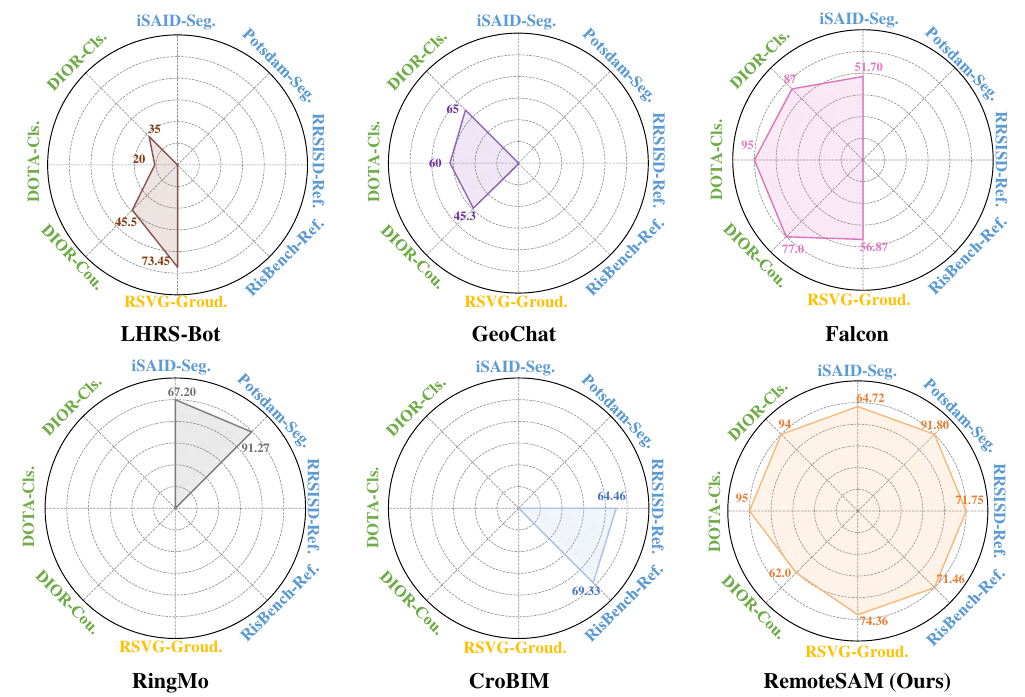}
  \caption{Comparison of various foundation models for Earth observation~\cite{muhtar2024lhrs,kuckreja2024geochat,yao2025falcon,sun2022ringmo,dong2024cross}. Blue, yellow, and green represent pixel-level, region-level, and image-level tasks, respectively. Our RemoteSAM is competitive with other models on most datasets and performs significantly better on pixel-level tasks.}
  \label{radar}
\end{figure}

Advances in AI have fundamentally transformed Earth observation paradigms~\cite{tuia2024artificial}. Strong visual perception models are driving breakthroughs across diverse applications, such as urban development~\cite{chen2024remote}, agriculture~\cite{victor2024remote}, disaster management~\cite{arnaudo2024fmars}, and beyond. As these tasks often involve distinct input-output interfaces, they are typically handled individually by specialized models. To effectively manage this heterogeneity, we are interested in developing a foundational model~\cite{zhang2024vision,awais2025foundation} that unifies multiple perception tasks. Such a model would conveniently accommodate varied application scenarios and effectively integrate knowledge learned across different tasks and domains.

\begin{figure*}[t]
  \centering
  \includegraphics[width=0.99\linewidth]{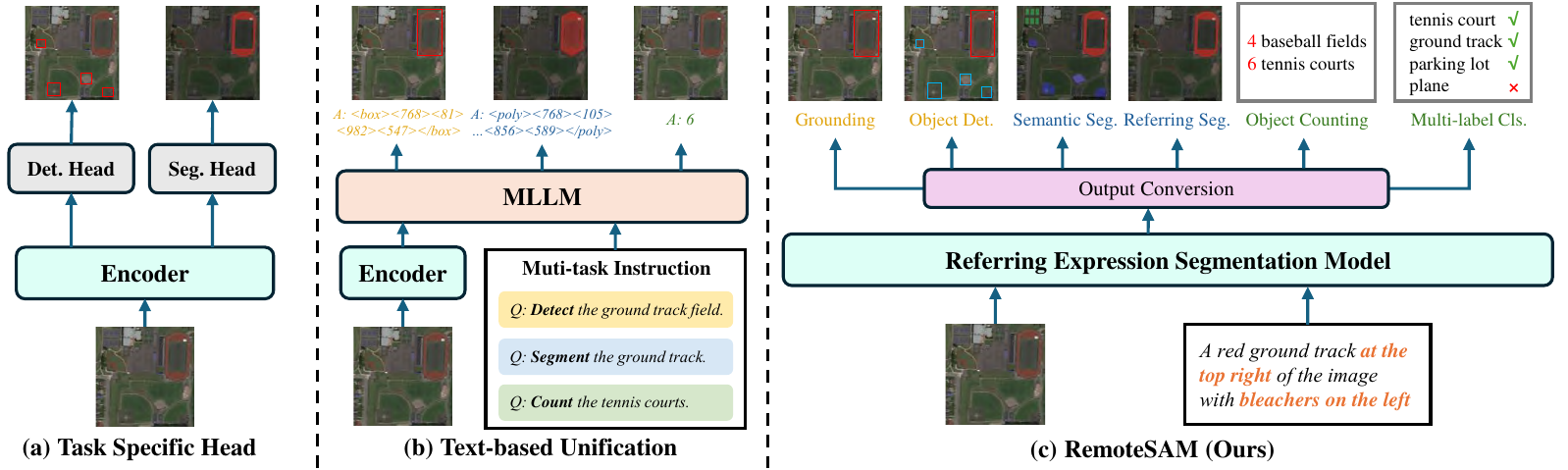}
  \caption{Comparison of different foundation models of remote sensing. (a) Task-specific head methods~\cite{cong2022satmae,li2024masked}. (b) Text-based unification approaches~\cite{kuckreja2024geochat,yao2025falcon}. (c) Our proposed referring segmentation-based paradigm, RemoteSAM. It is a more unified architecture with fewer parameters, achieving enhanced pixel-level understanding performance.}
  \label{fig1}
\end{figure*}



Current attempts at task unification mainly fall into two paradigms, as represented in Fig.~\ref{fig1}. Methods based on \textbf{task-specific heads} (\eg, Embedding Fields~\cite{brown2024learned}, RingMo~\cite{sun2022ringmo}, ScaleMAE~\cite{10377166}, and various extensions of RemoteCLIP models~\cite{liu2024remoteclip,zhang2024rs5m,li2024toward} ) integrate general-purpose encoders with specialized decoders tailored to individual downstream tasks.  However, the knowledge sharing between different task heads is inherently limited, and it necessitates task-specific fine-tuning whenever new objectives emerge.
Other methods, exemplified by Falcon~\cite{yao2025falcon}, GeoChat~\cite{kuckreja2024geochat}, employs \textbf{text-based task unification} with language models~\cite{chen2022unifieds,wang2022ofa,chen2022pix2seq}. Although achieving good performance in image-level (\eg, scene classification) and region-level tasks (\eg, object detection), such approaches exhibit intrinsic limitations in pixel-level prediction tasks, as natural language is not suitable for representing dense pixel-level outputs.



These limitations motivate us to design a flexible foundation model architecture capable of seamlessly supporting heterogeneous input-output interfaces across multiple tasks and granularities. Our proposed model operates at the fundamental pixel-level granularity, which acts as the most fundamental and indivisible output unit, enabling seamless upward compatibility to region-level and image-level tasks. Additionally, to facilitate convenient adaptation, the model integrates natural language understanding capabilities. This approach resembles the Referring Expression Segmentation (RES)~\cite{ji2024survey,hu2023beyond,wu2024towards,chng2024mask} task, where a model generates a pixel-level map from the input image conditioned on a text prompt.


In this paper, we present \textbf{RemoteSAM}, a vision foundation model built upon a novel architectural paradigm centered on RES. Our model leverages the dense pixel-level outputs inherent in RES, effectively converting these outputs into various formats required by other vision-centric tasks. Unlike existing VLM-based foundation models such as Falcon~\cite{yao2025falcon}, LHRS-Bot~\cite{muhtar2024lhrs}, EarthGPT~\cite{zhang2024earthgpt}, and GeoChat~\cite{kuckreja2024geochat}, our RemoteSAM seamlessly supports pixel-level (\eg,  segmentation), region-level (\eg,  grounding), and image-level (\eg,  counting) tasks within a unified architecture. Furthermore, by eliminating the large LLM backbone—which contributes minimally to visual perception—RemoteSAM achieves significant parameter efficiency, enabling it to efficiently process high-resolution remote sensing data.

To equip RemoteSAM with robust semantic understanding capabilities, an RES dataset with extensive semantic coverage is necessary. However, existing datasets~\cite{liu2024rotated,yuan2024rrsis} suffer from limited categorical diversity, restricted attribute variation, and overly templated expressions. To overcome these limitations, we construct a semantically diverse dataset using a scalable automated data curation pipeline. Observing that recent VLMs~\cite{qwen2025qwen25technicalreport} demonstrate strong image comprehension abilities, we leverage them to extract rich semantic information from remote sensing imagery, generating referring expressions with broad linguistic diversity. Through iterative pseudo-label refinement utilizing mixed teacher models, we create a comprehensive dataset containing 270K Image-Text-Mask triplets, named \textbf{RemoteSAM-270K}. It is characterized by two critical dimensions: 1) extensive category Scope, encompassing diverse and prevalent remote sensing targets; and 2) multifaceted attribute completeness, employing linguistically varied expressions to describe detailed object attributes such as colors, states, spatial relations, and other distinctive visual-semantic characteristics.

To quantitatively validate the semantic coverage of remote sensing datasets, we further establish a hierarchical remote sensing semantic vocabulary, named \textit{RSVocab-1K}, comprising 1K prevalent object categories. The qualitative results based on RSVocab-1K validate that RemoteSAM-270K's referring expressions possess a diverse range of category completeness and attribute expressiveness.
Benefiting from our high-quality dataset, the model achieves superior adaptability to more unseen categories and datasets.
We also conduct holistic evaluations to measure the performance of RemoteSAM across multiple downstream visual-centric tasks of remote sensing. 
The experiment result demonstrates that RemoteSAM effectively addresses the persistent performance gap of conventional foundation models in fine-grained pixel-wise prediction tasks with an order-of-magnitude smaller parameter count (from billions to millions). 
For example, RemoteSAM achieves significant performance improvements on referring expression segmentation, outperforming existing methods by more than 3.0\% on both RRSISD and RisBench benchmarks in terms of mIoU. It also achieves state-of-the-art semantic segmentation performance without fine-tuning, surpassing vision foundation models, \eg, MA3E~\cite{li2024masked} and ScaleMAE~\cite{reed2023scale}. In addition, it yields a remarkable 35\% accuracy gain over GeoChat~\cite{kuckreja2024geochat} in multi-label classification with a parameter-efficient architecture.

The contributions of this paper to remote sensing are summarized as follows:
\begin{itemize}

    \item We propose a robust yet flexible visual foundation model for Earth observation, RemoteSAM. To the best of our knowledge, it is the first exploration and practice of referring segmentation-based task unification paradigm.
    
    \item We build a new referring expression dataset with an automatic data curation pipeline. It significantly exceeds the scale of existing datasets and possesses rich semantic coverage, benefiting from the advantages of VLMs.
    \item We construct a hierarchical semantic vocabulary. It can be utilized to evaluate the semantic coverage of remote sensing datasets, which helps measure their ability to adapt to real-world applications.
    \item Holistic evaluations demonstrate that RemoteSAM's superior performance over existing approaches with prominent fewer parameters. It demonstrates remarkable performance across multiple visual-centric tasks, particularly in pixel-level interpretation tasks.
\end{itemize}

\section{Related Work}
\subsection{Remote Sensing Foundation Models}

Recent advances in foundation models~\cite{qwen2025qwen25technicalreport,awais2025foundation,liu2024few} have catalyzed transformative progress across multiple computer vision domains. However, despite their success in processing natural images, existing VFMs exhibit critical limitations when applied to remote sensing imagery~\cite{yao2024domain,liu2024scale,miao2025prompting} — a domain characterized by multimodal signals (\eg, multispectral bands), fine-grained spatial details (\eg, sub-meter resolutions), complex geospatial relationships, and temporal dynamics across acquisition epochs. 
Therefore, researchers have initiated systematic development of remote sensing foundation models.
Initially, RingMo~\cite{9844015} is the first to construct a generative self-supervised framework. Through multi-modal data augmentation and scene-aware contrastive learning, it solves the problem of representation in complex remote sensing scenes. 
To capture temporal evolution patterns, SatMAE~\cite{cong2022satmae} designs a temporal embedding encoding and cross-time mask reconstruction strategy, establishing a new paradigm for dynamic remote sensing sequence modeling. To address the bottleneck in multi-scale representation, Scale-MAE~\cite{10377166} proposes a scale-aware masking strategy and a hierarchical decoder architecture, achieving scale-invariant learning of geospatial features. BFM~\cite{Cha_2024} constructs a large remote sensing model with billions of parameters for the first time through a mixture-of-experts architecture and distributed training technology, verifying the feasibility of model scaling. The recently proposed SatMAE++~\cite{10657832} further integrates multi-resolution pre-training and a convolutional upsampling module, achieving hierarchical fusion of cross-scale features, thereby integrating multi-scale information and enhancing the modeling capability of remote sensing images. Inspired by MiniGPT-4~\cite{zhu2023minigpt}, GeoChat~\cite{kuckreja2024geochat} achieves the ability of visual grounding by utilizing a novel multimodal instruction dataset. LHRS-Bot~\cite{muhtar2024lhrs} creates LHRS-Instruct, using globally available remote sensing images and corresponding OpenStreetMap features to exhibit a profound comprehension of RS images.


\begin{figure*}[t]
  \centering
  \includegraphics[width=0.99\linewidth]{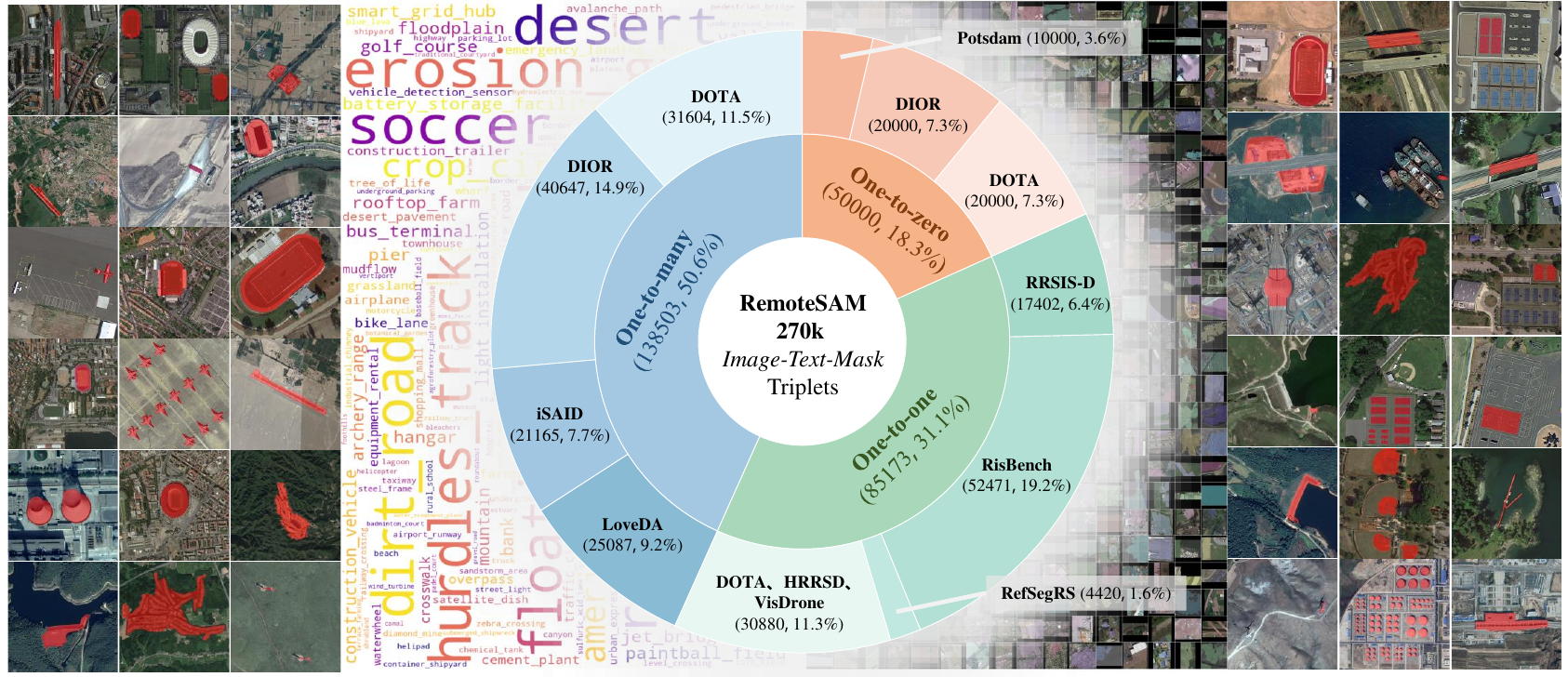}
  \caption{The composition of our proposed RemoteSAM-270K dataset. It is a generalized referring segmentation dataset, which is constructed through multi-source integration of mainstream remote sensing datasets, with semantically-dense referring expressions generated by VLMs.}
  \label{dataset}
\end{figure*}

\subsection{Remote Sensing Referring Segmentation}
Referring Image Segmentation~\cite{ji2024survey,chng2024mask} aims to segment specific objects from natural images utilizing natural language descriptions. Existing natural image segmentation methods are difficult to apply directly to remote sensing images. LGCE~\cite{yuan2024rrsis} defines the Referring Remote Sensing Image Segmentation (RRSIS) task and provides the RefSegRS dataset. It also presents a language-guided cross-scale enhancement module designed to improve small object segmentation by integrating multi-scale visual and linguistic features. However, the limited data scale constrains LGCE's performance with complex data. Consequently, RMSIN~\cite{liu2024rotated} proposed the RRSIS-D dataset and addresses the diverse scales and orientations of objects in remote sensing images through intra-scale and cross-scale interaction modules as well as adaptive rotated convolution.
To overcome the limitations of current RRSIS datasets—small size, single spatial resolution, and sparse object samples—CroBIM~\cite{dong2024cross} introduces RISBench, offering more comprehensive and challenging data. They also design a cross-modal bidirectional interaction model with context-aware prompt modulation, which enhances segmentation performance in complex remote sensing backgrounds. Considering that existing RRSIS methods typically employ a simple and direct image-text alignment approach, neglecting the fine-grained relationships between images and text descriptions, 
FIANet~\cite{lei2024exploring} proposes fine-grained image-text alignment and text-aware multi-scale enhancement modules. These modules improve the segmentation of targets in complex remote sensing scenes.

\section{RemoteSAM-270K Dataset}

\subsection{Improving Semantic Coverage of Data} 
Existing referring remote sensing image segmentation datasets exhibit limited semantic coverage, failing to support a foundation model's strong capabilities. To address this gap, we propose RemoteSAM-270K, a large-scale referring expression segmentation dataset that expands category coverage and attribute diversity for rich semantic coverage.

\subsubsection{Category Expansion} As shown in Fig.~\ref{dataset}, we propose to enrich the category completeness from existing remote sensing datasets. Specifically, we follow the steps outlined below to generate pixel-wise annotations.

\textbf{Datasets Integration:} We collect diverse remote sensing datasets (\eg, iSAID~\cite{waqas2019isaid}, LoveDA~\cite{wang2021loveda}, DOTA~\cite{xia2018dota}, HRRSD~\cite{zhang2019hierarchical}) and standardize their formats. If the dataset is region-level, such as HRRSD~\cite{zhang2019hierarchical}, DOTA~\cite{xia2018dota}, we employ the code of SAMRS~\cite{wang2023samrs} to generate corresponding instance-level masks via their detection annotations.

\textbf{Triplets Generation:} We construct \textit{image-text-mask} triplets for referring segmentation through three distinct data generation strategies, adhering to the generalized referring expression segmentation paradigm in benchmarks like G-RefCOCO~\cite{liu2023gres} and RefZOM~\cite{hu2023beyond}:
(1) ``\textit{One-to-One}'': Directly integrate existing referring segmentation annotations from RefSegRS~\cite{yuan2024rrsis}, RRSISD~\cite{liu2024rotated}, and RisBench~\cite{dong2024cross}.
(2) ``\textit{One-to-Many}'': Generate referring expressions following the template {``\{category\} in the image.''} and decompose original segmentation masks into class-specific sub-masks. Each sub-mask aggregates all instances belonging to its corresponding category.
(3) ``\textit{One-to-Zero}'': Create null masks (all-zero matrices) paired with textual descriptions of categories explicitly absent from the image.
This type of sample can effectively prevent the model from generating masks when the expressions refer to categories not present in the image.

\subsubsection{Attribute Expansion} 
Expressions with diverse attributes and flexible structures facilitate the model's capacity to learn rich semantic representations~\cite{chen2024makes}. To achieve this objective, we integrate an automatic referring expression generation method with a semi-supervised Mixed-Teacher framework, producing high-quality \textit{Image-Text-Mask} triplets enriched with detailed attributes.


\textbf{Expressions Creation:} Inspired by the Pyramid of Captions (PoCa) method~\cite{chen2024makes}, we split images into multiple local patches to expand the original remote sensing datasets (\eg, DOTA~\cite{xia2018dota}, HRRSD~\cite{zhang2019hierarchical}). Then, we employ Qwen2-VL-72B-Instruct~\cite{qwen2025qwen25technicalreport} with the prompt ``\textit{Describe all of your observations in this image as comprehensively as possible.}'' to generate attribute-enriched captions. 
However, these captions tend to focus on the information of the whole image rather than on a specific object. They cannot be directly utilized as referring expressions. Therefore, we design prompts (shown in Supplymentary) to further parse these captions into multiple descriptions, each containing only a single category target that has relevant referring information.  


\begin{table}[t]
\centering
\caption{Comparison of different referring remote sensing segmentation datasets. ``Generalized'': contains multi-target, no-target, and single-target expressions, ``Cls'': Categories, ``Attr'': Attributes.}
\renewcommand\arraystretch{1.05}
\resizebox{0.48\textwidth}{!}{
\begin{tabular}{c|ccccc}
\toprule
Dataset        & Generalized & \# Samples & \# Cls & \# Attr & \# Attr/Sample \\
\hline
RefSegRS~\cite{yuan2024rrsis}       & \ding{53}          & 4.4k    & 15         & 3            & 0.78                          \\
RRSIS-D~\cite{liu2024rotated}        & \ding{53}          & 17.4k   & 20          & 7            & 2.41                          \\
RISBench~\cite{dong2024cross}       & \ding{53}           & 52.5k   & 26          & 8            & 2.45                          \\
\hline  \rowcolor{blue!10}
\textbf{RemoteSAM-270K} & \checkmark          & \textbf{270K}   & \textbf{297}          & \textbf{16}           & \textbf{3.17}   \\ 
\bottomrule
\end{tabular}
}
\label{datasets}
\end{table}

\begin{figure}[t]
  \centering
  \includegraphics[width=0.99\linewidth]{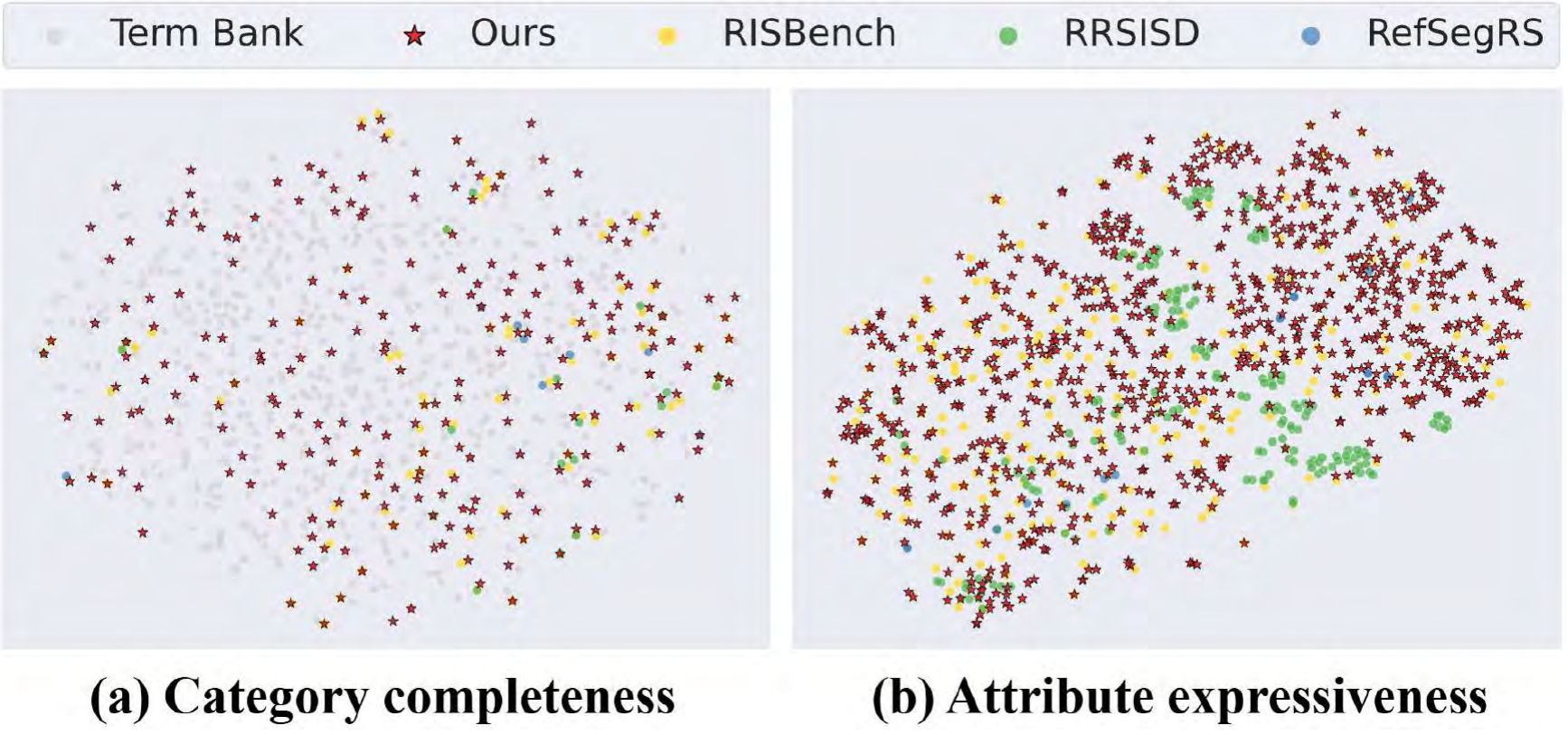}
  \caption{Comparison of semantic coverage on ours and other remote sensing referring image segmentation datasets.}
  \label{tsne}
\end{figure}

\textbf{Masks Production:} 
After obtaining the expressions, we also require a strategy for automatically generating masks. To achieve this objective, we utilize a mixture of several types of expert models (\eg, GroundedSAM2~\cite{ren2024grounded}, RMSIN~\cite{liu2024rotated}, etc.) to generate pseudo-labels. 
Unfortunately, even with multiple expert models, there are still a significant number of unreliable samples among the generated pseudo-labels. Therefore, to ensure data quality, we employ SigLIP2 \cite{tschannen2025siglip} to calculate the similarity between the mask-related image regions and the corresponding expressions, removing samples with low similarities. This process is iterated, ultimately yields a set of attribute-rich triplets.


\subsection{Data Analysis} 
We compare category and attribute distributions across four existing remote sensing referring image segmentation benchmarks. As illustrated in the Tab.~\ref{datasets}, our RemoteSAM-270K dataset contains over 270,000 triplets of \textit{image-text-mask}, spanning 297 categories and 16 types of fine-grained attribute descriptions. Furthermore, RemoteSAM-270K is the first generalized~\cite{hu2023beyond} remote sensing referring segmentation dataset, which can also enhance the depth of remote sensing referring segmentation research.

To facilitate the analysis of semantic coverage, we build a remote sensing vocabulary named RSBocab-1K. We integrate 1,000 fine-grained categories aligned with USGS Land Cover\footnote{https://www.usgs.gov/programs/gap-analysis-project/science/land-cover} and GB/T 21010-2017 remote sensing object classification specifications. Then, we organize these categories into three hierarchical levels.
Following RSVocab-1K, we visualize the category completeness and attribute expressiveness utilizing t-SNE projection in Fig.~\ref{tsne}. The visualization reveals two critical advantages of our RemoteSAM-270K: (1) superior category completeness and (2) richer attribute expressiveness. This enhanced comprehensiveness originates from our integration of cross-domain remote sensing datasets combined with VLMs that effectively mine latent semantic patterns.

\begin{figure}[t]
  \centering
  \includegraphics[width=0.99\linewidth]{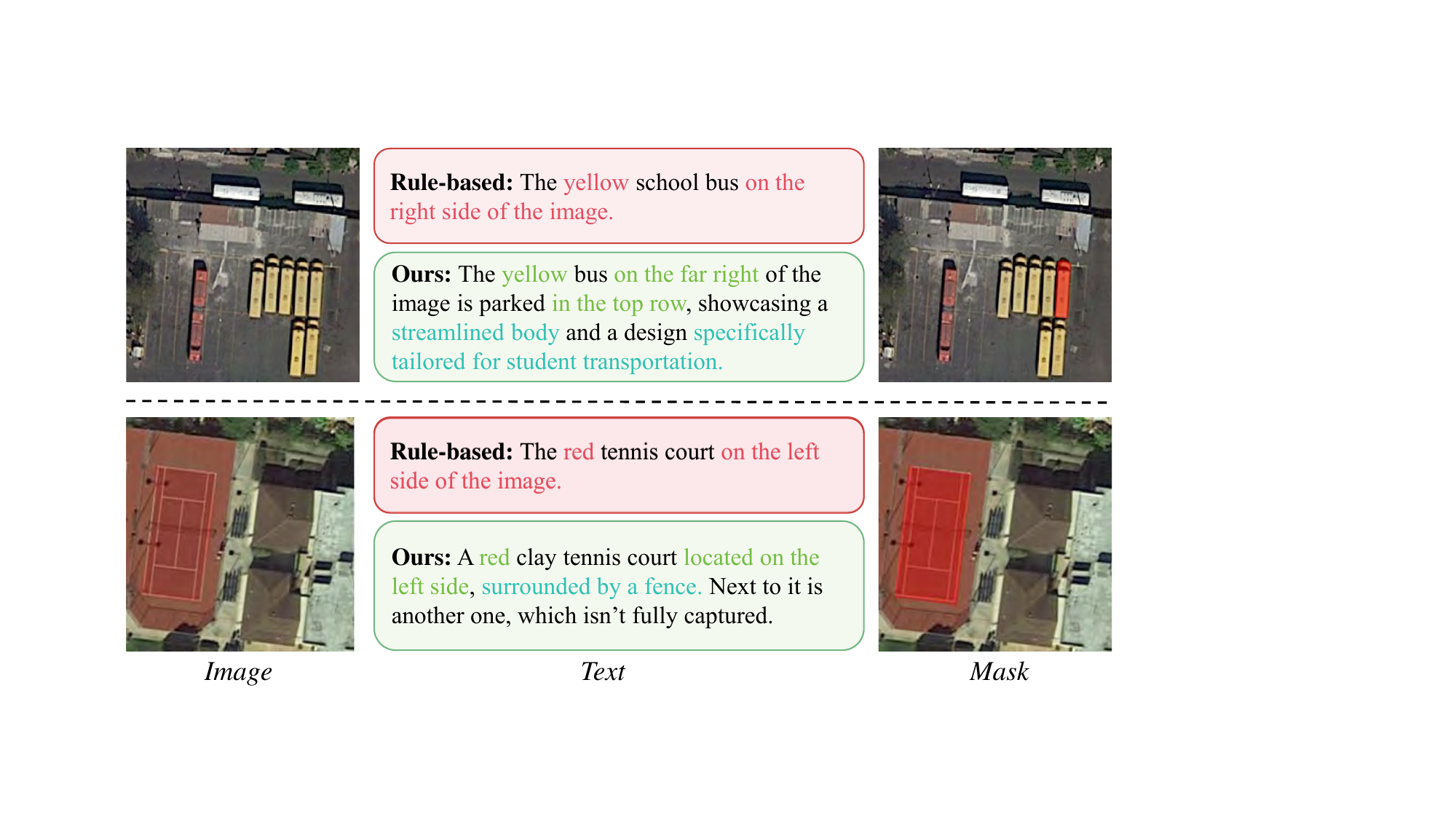}
  \caption{Comparison between VLMs (ours) and rule-based (RRSISD) generated expressions.}
  \label{lvlms}
\end{figure}

In Fig.~\ref{lvlms}, we showcase the advantages of referring expressions generated by the VLMs. It is evident that the rule-based generated expressions tend to have a more uniform structure, sometimes leading to ambiguities. For instance, the reference to ``\textit{the yellow school bus on the right side}'' is not clear. In contrast, the expressions generated by VLMs include more detailed attribute descriptions and adhere to a more flexible syntax, which can help the model to learn more complex semantic information.

\begin{figure*}[t]
  \centering
  \includegraphics[width=0.99\linewidth]{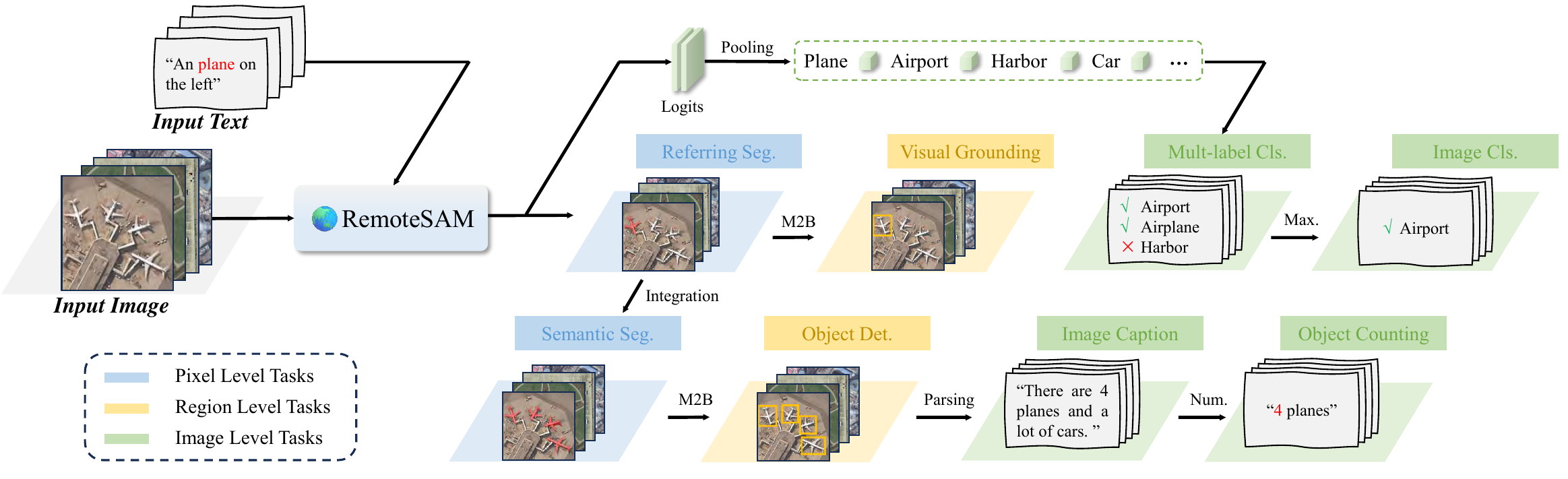}
  \caption{Overview of our proposed RemoteSAM. It is a foundational model centered around a referring expression segmentation framework. This generalist vision system demonstrates multi-task visual understanding capabilities through a single unified architecture without task-specific heads.}
  \label{overview}
\end{figure*}

\section{RemoteSAM}


Pixel-level mask serves as the foundational computation unit in vision tasks, ensuring native compatibility with higher-level tasks at both region and image scales while preserving maximum spatial precision.
Motivated by this insight, we build RemoteSAM through a unified task transition framework leveraging RES outputs through these strategies, as presented in Fig.~\ref{overview}. 

Formally, given a training dataset $D_{\text{train}} = \{(I_k, T_k, M_k)\}_{k=1}^N$ composed of $N$ \textit{Image-Text-Mask} triplets, where each image $I_k \in \mathbb{R}^{H \times W \times 3}$ is paired with a referring expression $T_k = \{t_1, \dots, t_n\}$ and its corresponding binary ground-truth mask $M_k \in \{0,1\}^{H \times W}$, RemoteSAM aims to predict a segmentation mask $\hat{M}$ for a query pair $(I, T)$ at inference. The prediction is formulated as $\hat{M} = \mathcal{Q}(\mathcal{F}_v(I), \mathcal{F}_t(T))$, where $\mathcal{F}_v: \mathbb{R}^{H \times W \times 3} \to \mathbb{R}^{H' \times W' \times D}$ and $\mathcal{F}_t: \{w_i\}_{i=1}^n \to \mathbb{R}^C$ denote visual and textual encoders respectively, and $\mathcal{Q}$ is a fusion-decoder that jointly reasons over cross-modal features to generate pixel-wise predictions. 
The model is trained by minimizing a segmentation loss $\mathcal{L}_{\text{seg}}$ between $\hat{M}$ and $M$.

Building upon the base prediction $\hat{M}$, we introduce a multi-task conversion strategy that transforms $\hat{M}$ into outputs for other vision tasks via task-specific functions $\{T_i: \hat{M} \to Y_i\}$, where $i$ denotes the $i^{th}$ task, $Y_i$ defines the output space of task (\eg, bounding boxes for detection, class probabilities for multi-label classification). Each task and its specific implementation steps are as follows.

\subsection{Pixel-level Tasks} 

\textbf{Referring Segmentation:} Referring expression segmentation aims to segment specific objects or regions in an image via a free-form referring expression. This task serves as the foundational task for RemoteSAM and can directly utilize the model's original output mask $\hat{M}$.

\textbf{Semantic Segmentation:} Since we can obtain segmentation masks of any instance via generalized referring expressions, straightforwardly integrating all masks can produce a semantic segmentation result.
Given a category set $C$ for the specific segmentation task, for each class $c \in C$ we generate referring expression $t_c$ = ``All $\{c\}$ in the image'' to acquire all the masks of this category:
\begin{equation}
\hat{M}_c = \mathcal{Q}\left(\mathcal{F}_v(I), \mathcal{F}_t(t_c)\right) \odot  \left(P(c|I) \geq \tau_{\text{seg}}\right),
\label{eq:semseg}
\end{equation}
where $\hat{M}_c \in \{0,1\}^{H \times W}$ is the predicted mask, $P(c|I)$ is category confidence with $\tau_{\text{seg}}$ as threshold. Then, we can iteratively process all categories through the referring segmentation model, aggregating instance masks into semantic segmentation maps.

\subsection{Region-level Tasks} 

\textbf{Visual Grounding:} Visual Grounding can be regarded as a type of ``referring expression detection'' task. Therefore, we directly convert predicted segmentation masks into bounding boxes via mask-to-bbox (M2B)~\cite{liu2024remoteclip} method $\mathcal{F}_{M2B}$.
Specifically, for each predicted referring mask $\hat{M} \in \{0,1\}^{H \times W}$, compute grounding bounding box $B$ coordinates:
\begin{equation}
B =\mathcal{F}_{M2B}(\hat{M}) = \left[(\min x_i,\ 
                \min y_i), (\max x_j,\ 
                \max y_j)\right].
\label{m2b}
\end{equation}

\textbf{Object Detection:} It shares conceptual similarities with semantic segmentation in visual recognition tasks. The primary distinction lies in their annotation formats: region-level bounding boxes and pixel-level masks. Therefore, semantic segmentation masks can be converted into rectangular bounding boxes. However, the M2B strategy encounters limitations when processing adjacent targets. Specifically, the overlapping regions of neighboring masks tend to merge during segmentation, resulting in a unified bounding box that encapsulates multiple distinct objects. To prevent adjacent objects from merging, we employ EPOC~\cite{chen2024subobject} to detect the objects' contours and refine mask boundaries. Then, we can utilize the M2B strategy (Eq.~\ref{m2b}) to convert the semantic segmentation outputs into multiple bounding boxes. 



\subsection{Image-level Tasks} 

\textbf{Multi-Label Classification:} Since semantic segmentation maps indicate the existence of each category, intuitively, they can be converted into classification results.
Specifically, we compute confidence scores through pooling over spatial-weighted probability maps of semantic segmentation outputs, followed by class-wise probability aggregation:
\begin{equation}
S_c = \lambda\cdot{{\frac{1}{H W}}\sum_{i,j}P(c|i,j)}+(1-\lambda)\cdot{\operatorname*{max}_{i,j}P(c|i,j)}, 
\end{equation}
where $\lambda$ denotes balance parameters of max and average pooling, $P(c|i,j)$ represents the normalized probability of class $c$ in position $(i,j)$ of the input image.

Then labels of class $c$ are judged as positive when their corresponding confidence score $S_c$ exceeds confidence threshold $\tau_{\mathrm{cls}}$ (0.5 as default):
\begin{equation}
y_{multi} = \{S_k \geq\tau_{\mathrm{cls}}\}.  
\end{equation}

\textbf{Image Classification:} Confidence scores are obtained in the same manner as multi-label classification, with the distinction being that we select the class with the highest confidence score:

\begin{equation}
y_{scene} = \arg\max \left[S_1,\ldots,S_C\right].
\end{equation}

\begin{table}[t]
\centering
\caption{Comparison of semantic segmentation results on unseen dataset with various open-vocabulary and referring segmentation models.}
\resizebox{0.48\textwidth}{!}{
\begin{tabular}{l|c|ccc}
\toprule
\multirow{2}{*}{Methods} & \multirow{2}{*}{Publication} & Vaihingen & UDD5   & DeepGlobe    \\
\cline{3-5}
                         &                              & $mIoU$ (\%) & $mIoU$ (\%) & $mIoU$ (\%) \\
\hline  \rowcolor{gray!20}
\multicolumn{2}{l|}{\textit{Open-vocabulary Models}}                            & & & \\
            
MaskCLIP~\cite{zhou2022maskclip}                 & ECCV22                       & 24.7      & 32.4     & 13.2\\
SCLIP~\cite{wang2023sclip}                    & arXiv22                      & 28.4      & 38.7     & 7.0 \\
GEM~\cite{Bousselham_2024}                      & CVPR24                       & 24.7      & 41.2     & 4.7 \\
ClearCLIP~\cite{lan2024clearclip}                & ECCV24                       & 27.3      & 41.8    & 5.7 \\
SegEarth-OV~\cite{li2024segearth}              & arXiv24                      & \underline{29.1}      & \textbf{50.6}    & 20.1 \\
\hline  \rowcolor{gray!20}
\multicolumn{2}{l|}{\textit{Referring Models}}                            & & &\\
FIANet~\cite{lei2024exploring}              & TGRS24                      &    1.1   & 2.4   & 47.8  \\
RMSIN~\cite{liu2024rotated}              & CVPR24                      &  2.2      &  1.7  &  \underline{47.9} \\
\hline \rowcolor{blue!10}
\textbf{RemoteSAM}                     & -                            & \textbf{46.0}      & \underline{45.6}    &  \textbf{60.5}\\
\bottomrule
\end{tabular}
}
\label{unseen}
\end{table}

\begin{figure}[t]
  \centering
  \includegraphics[width=\linewidth]{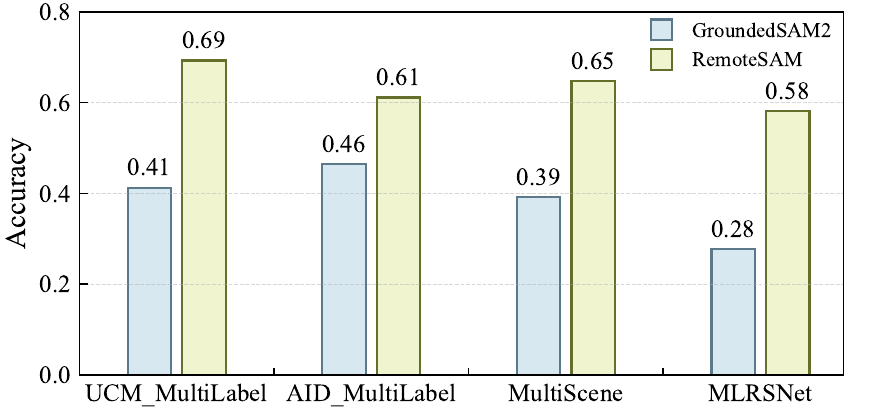}
  \caption{Comparison of zero-shot classification on SATIN with Grounded SAM2.}
  \label{satin}
\end{figure}

\textbf{Image Caption:} When we possess object positions $B$, categories $y_{multi}$, and numbers $N_c$ of objects, it is possible to generate captions in a rule-based manner~\cite{liu2024remoteclip,zhan2023rsvg}. Specifically, $y_{multi}$ provides an overview of image categories, while $N_c$ and spatial relationships from $B$ yield finer details.


\textbf{Object Counting:} As we can detect objects in the image, counting the number of class $c_{target}$ can be easily performed:
\begin{equation}
N_c = \sum_{i=1}^V \mathbb{I}(c_i = c_{\text{target}}),
\end{equation}
where $\mathbb{I}$ denotes the indicator function, $V$ is the total number of detected bboxes, $c_i$ is the $i^{th}$ bbox's category.


\section{Experiments}
In this section, we conducted comprehensive experiments to evaluate the semantic coverage of RemoteSAM-270K and the task unification of RemoteSAM. 
Semantic coverage evaluates the generalization performance of the model trained on our dataset, including segmentation on unseen categories and zero-shot classification. Task unification evaluation is realized by testing the performance of various downstream visual-centric tasks, including pixel-level, region-level, and image-level. Additional experimental results are provided in the supplementary materials.

\subsection{Evaluation Setup}
\subsubsection{Models}  Our implementation built upon the RMSIN~\cite{liu2024rotated} architecture with BERT~\cite{devlin2019bert} as the textual encoder and Swin-Base~\cite{liu2021swin} as the visual encoder. For downstream tasks, we compared three categories of candidates: (1) Vision pre-trained foundation models (VFMs): RingMo~\cite{sun2022ringmo}, ScaleMAE~\cite{reed2023scale}, MA3E~\cite{li2024masked}, et al; (2) Vision-language models (VLMs): GeoChat~\cite{kuckreja2024geochat}, LHRS-Bot~\cite{muhtar2024lhrs}, Falcon~\cite{yao2025falcon}, et al; (3) Task-specific architectures: SegEarth-OV~\cite{li2024segearth}, GeoGround~\cite{zhou2024geoground}, MGVLF~\cite{zhan2023rsvg}, et al.
\subsubsection{Datasets} We compared our approach against previous SOTA referring segmentation models across two large-scale datasets: RRSISD~\cite{liu2024rotated} and RisBench~\cite{dong2024cross}. The downstream evaluation protocol spans five established benchmarks in remote sensing: DOTA~\cite{xia2018dota}, DIOR~\cite{li2020dior}, iSAID~\cite{waqas2019isaid}, Potsdam\footnote{https://www.isprs.org/education/benchmarks/UrbanSemLab/2d-sem-label-potsdam.aspx}, and RSVG~\cite{zhan2023rsvg}.
\subsubsection{Training Settings.}  The experimental configuration employed 8 × NVIDIA GeForce RTX 4090 GPUs with image resolution fixed at 896×896. Training proceeds for 40 epochs using AdamW optimization, initialized with a learning rate of 3e-5.

\begin{table}[t]
\centering
\caption{Comparison of referring expression segmentation results on RRSISD and RisBench.}
\renewcommand\arraystretch{1.1}
\resizebox{0.48\textwidth}{!}{
\begin{tabular}{l|c|cc|cc}
\toprule
\multirow{2}{*}{Methods} & \multirow{2}{*}{Publication} & \multicolumn{2}{c|}{RRSISD}                                                                                 & \multicolumn{2}{c}{RisBench}                                                                               \\ \cline{3-6}

                         &                              &  $oIoU (\%)$ & $mIoU (\%)$ & $oIoU (\%)$ & $mIoU (\%)$ \\
\hline
BRINet~\cite{hu2020bi}                   & CVPR20                                & 69.88         & 49.65          & 48.73         & 42.91      \\
LSCM~\cite{hui2020linguistic}                      & ECCV20                                & 69.05         & 49.92        & 50.08         & 43.69      \\
CMPC~\cite{huang2020referring}                      & CVPR20                          & 69.22         & 49.24          & 50.24         & 43.82      \\
CMPC+~\cite{liu2021cross}  & TPAMI21  & 70.13 & 50.12 & 53.98 & 46.73 \\
LAVT~\cite{yang2022lavt}                      & CVPR22                               & 77.19         & 61.04       & 74.15         & 61.93      \\
CroosVLT~\cite{cho2023cross}        & TMM23  & 75.48 & 58.48 & 74.33 & 62.84 \\
CARIS~\cite{liu2023caris}           & MM23  & 77.17 & 62.12 & 75.10 & 65.79\\
LGCE~\cite{yuan2024rrsis}                      & TGRS24                               & 76.34         & 59.37       & 73.87         & 62.13      \\
CroBIM~\cite{dong2024cross}                    & TGRS24                                & 75.99         & 64.46       & 73.04         & 68.03      \\
robust-ref-seg~\cite{wu2024towards}  & TIP24  & 77.40  &58.91 & 74.23 & 61.25 \\
RMSIN~\cite{liu2024rotated}                     & CVPR24                               & 77.88         & 64.26       & \underline{75.24}         & \underline{68.25}      \\

RS2-SAM2~\cite{rong2025customized}                  & arxiv25                        & \underline{78.99}       & \underline{66.72}       & -             & -          \\
\hline  \rowcolor{blue!10}
\textbf{RemoteSAM}            & -           & \textbf{80.04} & \textbf{71.75}  & \textbf{75.93} & \textbf{71.46}      \\
\bottomrule
\end{tabular}
}
\label{rrsis}
\end{table}

\subsection{Evaluations of Semantic Coverage}
In this section, we validated the semantic coverage of RemoteSAM-270K. We believe that the improvement in semantic coverage should lead to a qualitative enhancement in model performance, which can be demonstrated through the model's generalization abilities.
Firstly, we evaluated our RemoteSAM on unseen datasets, comparing it against several state-of-the-art open-vocabulary segmentation methods and other referring segmentation approaches, as illustrated in the Tab.~\ref{unseen}. 
On the Vaihingen dataset, RemoteSAM outperformed SegEarth-OV by 16.9\%, while other referring segmentation methods demonstrated significantly lower accuracy. These results indicate that our dataset effectively enhances the model's cross-domain generalization competence alongside demonstrated open-set identification capacity.

To further validate the enhancement of category diversity in our dataset, we also conducted experiments on zero-shot classification utilizing the SATIN~\cite{roberts2023satin} meta dataset, as represented in Fig.~\ref{satin}. SATIN contains over 250 categories, comprised of images with various resolutions and viewpoints. The results indicate that the model trained on our RemoteSAM-270K outperforms the Grounded SAM2~\cite{ren2024grounded} in recognition performance, which demonstrates that our data can effectively support the model in recognizing a majority of common remote sensing categories.


\begin{table}[t]
\centering
\caption{Comparison of semantic segmentation results with various vision foundation models. }
\renewcommand\arraystretch{1.1}
\resizebox{0.48\textwidth}{!}{
\begin{tabular}{l|c|c|cc}
\toprule
\multirow{2}{*}{Methods} & \multirow{2}{*}{Publication} & Pre-trained & iSAID & Potsdam \\ \cline{4-5}
                   &      &                          Data    & $mIoU$ (\%)  & $mF1$ (\%)    \\
\hline
SeCo~\cite{manas2021seasonal}                      & ICCV21      &  Sentinel-2~\cite{phiri2020sentinel}              & 57.20 & 89.03   \\
GASSL~\cite{yang2021graph}                     & NIPS21       &  MillionAID~\cite{long2021creating}            & 65.95 & 91.27   \\
SatMAE~\cite{cong2022satmae}                    & NIPS22      & fMoW~\cite{christie2018functional}                 & 62.97 & 90.63   \\
RingMo~\cite{sun2022ringmo}                    & TGRS22     & About 2M~\cite{sun2022ringmo}                  & \textbf{67.20} & 91.27   \\
RVSA~\cite{wang2022advancing}                      & TGRS22      &  MillionAID~\cite{long2021creating}               & 64.49 & -       \\
SSL4EO~\cite{wang2023ssl4eo}                    & GRSM23        &    SSL4EO-S12~\cite{wang2023ssl4eo}           & 64.01 & 91.54   \\
CACo~\cite{mall2023change}                      & CVPR23         &  MillionAID~\cite{long2021creating}               & 64.32 & 91.35   \\
SAMRS~\cite{wang2023samrs}                     & NIPS23      &  MillionAID~\cite{long2021creating}                  & 66.26 & 91.43   \\
ScaleMAE~\cite{reed2023scale}                  & ICCV23     & fMoW~\cite{christie2018functional}                    & 65.77 & 91.54   \\
RSCoTr~\cite{li2024rscotr}                    & TGRS24        & ImageNet-22k~\cite{deng2009imagenet}                 & -     & 90.67   \\
MA3E~\cite{li2024masked}                      & ECCV24       &  MillionAID~\cite{long2021creating}                  & 64.06 & 91.50   \\
\hline  \rowcolor{blue!10}
\textbf{RemoteSAM}                 & -                  &   RemoteSAM-270K           & 64.72      & \underline{91.80}       \\ 
\hline  \rowcolor{blue!10}
\textbf{RemoteSAM-FT}                  & -              &  RemoteSAM-270K               & \underline{67.01}     &  \textbf{93.54}      \\ 
\bottomrule
\end{tabular}
}
\label{semantic_seg}
\end{table}

\begin{table}[t]
\centering
\caption{Comparison of visual grounding performance with specialized and foundation models.}
\resizebox{0.45\textwidth}{!}{
\begin{tabular}{l|c|c|cc}
\toprule
\multirow{2}{*}{Methods} & \multirow{2}{*}{Publication} & \multirow{2}{*}{Parameters} &\multicolumn{2}{c}{RSVG}  \\ \cline{4-5}
                         &                 &             & $AP_{50} (\%)$  & $mIoU (\%)$               \\
\hline  \rowcolor{gray!20}
\multicolumn{2}{l|}{\textit{Specialized Models}}                  &           && \\

ZSGNet~\cite{sadhu2019zero}	                 &        ICCV19              &  140M      & 51.67	        & 44.12              \\
FAOA~\cite{liao2020real}	                 &      CVPR20   &          150M           & 70.86         & 62.86              \\
ResC~\cite{huang2021look}	                 &      CVPR21   &      150M                & 72.71	        & 64.24              \\
LBYL-Net~\cite{liao2022progressive}                 &     TIP22    &     155M                & 73.78	        & 65.92              \\
TransVG~\cite{li2021referring}	                 &    NIPS21  &       136M                  & 72.41	        & 63.56              \\
VLTVG~\cite{ye2022shifting}	                 &   CVPR22  &         155M                & 75.79	        & 66.32              \\
MGVLF~\cite{zhan2023rsvg}	                 &   TGRS23   &       136M                 & 76.78	        & 68.04             \\
GeoGround~\cite{zhou2024geoground}	         &     arxiv24  &       7B                 & 77.73	        &   -                 \\

\hline  \rowcolor{gray!20}
\multicolumn{2}{l|}{\textit{Foundation Models}}                            & && \\

MiniGPT-v2~\cite{zhu2023minigpt}               & arXiv23     & 7B                 & 46.64         &-                    \\
Qwen-VL-Chat~\cite{qwen2025qwen25technicalreport}             & arXiv23    & 7B                   & 44.76         &-                    \\
SkyEyeGPT~\cite{zhan2025skyeyegpt}                  & NIPS22       & 7B                  & 70.50         &-                    \\
LHRS-Bot~\cite{muhtar2024lhrs}                  & ECCV24       & 7B                & \underline{73.45}         & -                   \\
Falcon~\cite{yao2025falcon}                   & arXiv25     & 0.7B                  & 56.87        & -                   \\
\hline  \rowcolor{blue!10}
\textbf{RemoteSAM}                     & -                 &    \textbf{180M}       & \textbf{74.36}      & 65.07       \\ 
\bottomrule
\end{tabular}
}
\label{VG}
\end{table}



\subsection{Evaluations of Task Unification}
We further conducted a series of Downstream evaluations to measure the RemoteSAM's performance on three categories of downstream vision-centric tasks: pixel-level, region-level, and image-level tasks. Due to space limitations, we presented semantic segmentation, visual localization, multi-label classification, and object counting as the corresponding tasks for evaluation. 

\textbf{Performance on Referring Expression Segmentation.}
As illustrated in Tab.~\ref{rrsis}, we evaluated referring segmentation performance on remote sensing imagery. The RemoteSAM demonstrates substantial superiority over comparative approaches. Specifically, it achieves 71.75\% mIoU on RRSISD - surpassing the SAM2-based architecture (RS2-SAM2~\cite{rong2025customized}) by 5.03\%. Our approach also establishes a new state-of-the-art result on RisBench with 3.21\% absolute performance gain. This advancement primarily stems from the extensive semantic coverage in our training dataset. 

\textbf{Performance on Semantic Segmentation.} To validate the performance of RemoteSAM on the semantic segmentation task, we selected several vision pre-trained models for comparison. As presented in Tab.~\ref{semantic_seg}, the results indicate that our approach performs comparably to specialized vision backbone models without task-specific tuning and even attains state-of-the-art accuracy on the Potsdam dataset (91.80\%). Moreover, further performance improvements (67.01\% and 93.54\%) can be achieved by training the model on the specific dataset.

\begin{figure*}[t]
  \centering
  \includegraphics[width=0.99\linewidth]{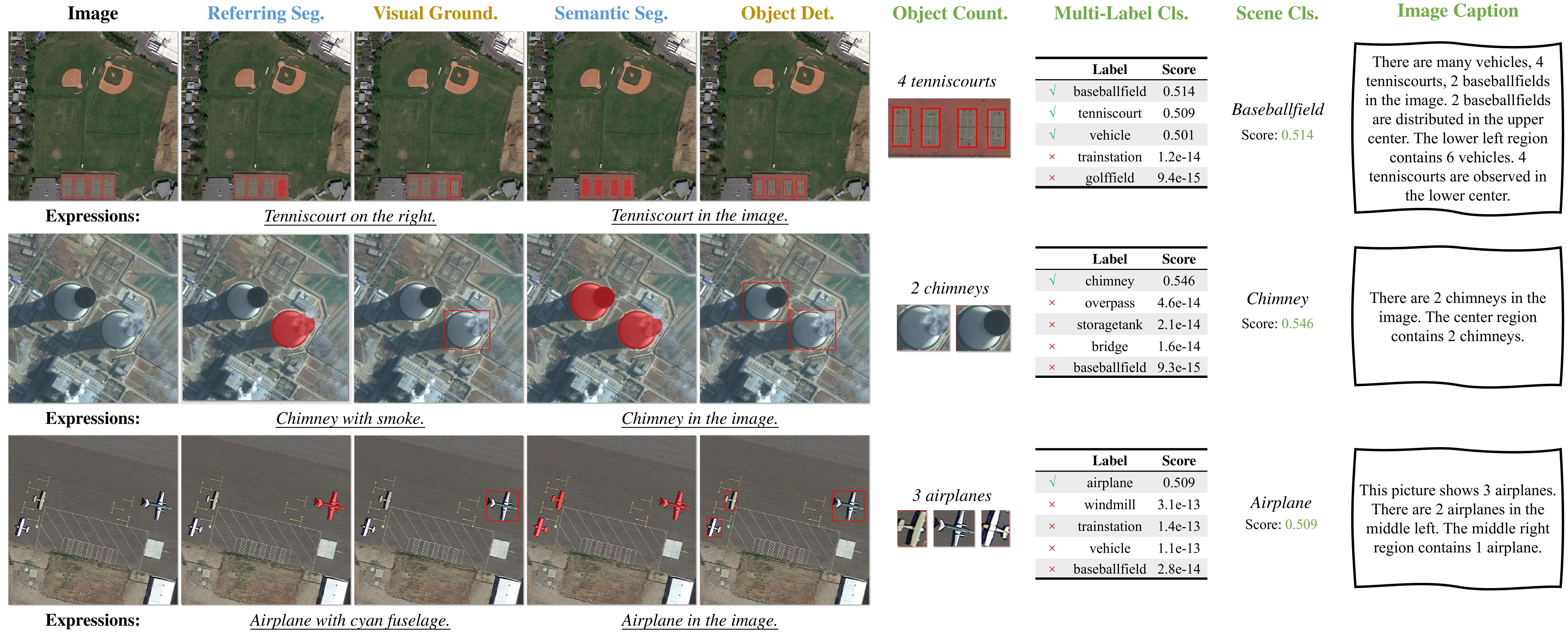}
  \caption{Inference examples of RemoteSAM on 8 visual-centric tasks.}
  \label{Visualization}
\end{figure*}

\begin{table}[t]
\centering
\caption{Comparison of multi-label classification results with various remote sensing foundation models.}
\resizebox{0.48\textwidth}{!}{
\begin{tabular}{l|c|c|cc}
\toprule
\multirow{2}{*}{Methods} & \multirow{2}{*}{Publication} & \multirow{2}{*}{{Parameters}} &  DIOR     & DOTAv2 \\ \cline{4-5}
                         &                           &   
                         & $Acc$     & $Acc$ \\
\hline
MiniCPM-V~\cite{hu2024minicpm}                 & arXiv24     &3B                 & 0.08     & 0.10     \\
MiniGPT-v2~\cite{zhu2023minigpt}                & arXiv23       &7B               & 0.16     & 0.11     \\
Qwen-VL-Chat~\cite{qwen2025qwen25technicalreport}              & arXiv23      &7B                & 0.17     & 0.10     \\
Sphinx~\cite{lin2023sphinx}                    & arXiv23        &7B              & 0.19     & 0.13     \\
LLaVA-1.5~\cite{liu2023llava}                 & NIPS24          &7B             & 0.35     & 0.19     \\
RemoteCLIP~\cite{liu2024remoteclip}                & TGRS24          &390M             & 0.59     & 0.46     \\
GeoChat~\cite{kuckreja2024geochat}                   & CVPR24        &7B               & 0.65     & \underline{0.60}     \\
LHRS-Bot~\cite{muhtar2024lhrs}                  & ECCV24        &7B               & 0.35     & 0.20     \\
Falcon~\cite{yao2025falcon}                  & arXiv25        &7B               & \underline{0.87}     & \textbf{0.95}     \\
\hline  \rowcolor{blue!10}
\textbf{RemoteSAM}         & -    &   \textbf{180M}   & \textbf{0.94} & \textbf{0.95} \\
\bottomrule
\end{tabular}
}
\label{cls}
\end{table}

\textbf{Performance on Visual Grounding.} Beyond pixel-level tasks, our RemoteSAM also supports fine-grained region-level tasks. For example, we present the visual grounding performance of RemoteSAM on the RSVG~\cite{zhan2023rsvg} dataset, comparing it against specialized and foundation models. As illustrated in Tab.~\ref{VG}, RemoteSAM performed as accurately as specialized models. Compared to other VLMs, our approach outperformed them significantly while utilizing a substantially smaller number of parameters.

\textbf{Performance on Multi-label Classification.} We evaluated 
the performance of RemoteSAM and VLMs on multi-label classification, with the results presented in Tab.~\ref{cls}. In the multi-label classification task, RemoteSAM achieved accuracy rates of 94\% and 95\% on the DIOR and DOTA datasets, surpassing GeoChat by 29\% and 35\%, respectively. This improvement may stem from the model's remarkable capability to comprehend complex spatial relationships in remote sensing scenes compared to VLMs.

\begin{table}[t]
\centering
\caption{Comparison of object counting results with various remote sensing foundation models.}
\resizebox{0.48\textwidth}{!}{
\begin{tabular}{l|c|c|cc}
\toprule
\multirow{2}{*}{Methods} & \multirow{2}{*}{Publication} & \multirow{2}{*}{{Parameters}} &  DIOR     & DOTAv2 \\ \cline{4-5}
                         &                           &   
                         & $Acc$     & $Acc$ \\
\hline
MiniCPM-V~\cite{hu2024minicpm}                 & arXiv24     &3B                 & 0.426 & \underline{0.260}     \\
MiniGPT-v2~\cite{zhu2023minigpt}                & arXiv23       &7B              & 0.429 & 0.248     \\
Sphinx~\cite{lin2023sphinx}                    & arXiv23        &7B               & 0.430 & 0.257     \\
LLaVA-1.5~\cite{liu2023llava}                 & NIPS24          &7B             & 0.249 & 0.221    \\
GeoChat~\cite{kuckreja2024geochat}                   & CVPR24        &7B              & 0.453 & 0.240    \\
LHRS-Bot~\cite{muhtar2024lhrs}                  & ECCV24        &7B               & \underline{0.455} & 0.244     \\
\hline  \rowcolor{blue!10}
\textbf{RemoteSAM}         & -    &   \textbf{180M}  & \textbf{0.620} & \textbf{0.409} \\
\bottomrule
\end{tabular}
}
\vspace{-0.25cm}
\label{count}
\end{table}

\textbf{Performance on Object Counting.} To further validate the performance of RemoteSAM at the image level, we conducted experiments on the object counting task. The task requires the model's ability to perceive and reason compositionally. As presented in Tab.~\ref{count}, our RemoteSAM attained accuracy rates of 62.0\% and 40.9\%, significantly outperforming other approaches such as LHRS-Bot. 

The above results demonstrate that our referring segmentation-centered visual unification paradigm achieves remarkable performance across pixel-level, region-level, and image-level tasks. It highlights RemoteSAM's ability to handle complex remote sensing tasks. More importantly, RemoteSAM has a significantly smaller number of parameters compared to other large-scale vision-language models and does not require task-specific decoders, enabling it to efficiently process high-resolution remote sensing images. 




\subsection{Further Analysis}
We presented a qualitative analysis of RemoteSAM across eight visual tasks in Fig.~\ref{Visualization}. Overall, RemoteSAM successfully translates the results of referring expression segmentation into outputs required for various other visual-centric tasks with precision, ranging from pixel-level to image-level. From a variety of examples, it is evident that RemoteSAM is capable of (1) localizing the boundaries of specified targets accurately (referring segmentation, semantic segmentation); (2) analyzing the relationships between objects in images (object detection, visual grounding); and (3) understanding the global contextual information in images (counting, classification, captioning). Specifically, in addition to locating objects via attributes such as location (first row), RemoteSAM also comprehends status information. For example, in the second row, it identifies a chimney that is smoking. Moreover, when provided with a generalized `one-to-many' expression (\eg, “Airplane in the image.”), RemoteSAM can further execute additional visual tasks.



\section{Conclusion}

In this work, we present \textbf{RemoteSAM}, a unified visual foundation model for Earth observation that addresses the critical pixel-level limitations of existing remote sensing foundation models through a referring segmentation-based paradigm. 
By developing an automated data curation pipeline leveraging VLMs and multi-teacher localization, we construct the largest referring expression segmentation dataset, RemoteSAM-270K (270K \textit{image-text-mask} triplets), with significant semantic diversity spanning 297 categories and 16 attributes. We also build a hierarchical semantic vocabulary to evaluate the semantic coverage of remote sensing datasets. 
Extensive evaluations demonstrate RemoteSAM's superiority in handling classification, detection, segmentation, and grounding with significant parameter efficiency. 
Our work demonstrates that segmentation-centric architectures can serve as unified backbones for multimodal Earth observation intelligence.

\section*{Acknowledge}

This work was supported in part by the National Natural Science Foundation of China under Grant 62372155 and Grant 62302149, in part by the Aeronautical Science Fund under Grant 2022Z071108001, in part by the Qinglan Project of Jiangsu Province, and in part by Changzhou Science and Technology Bureau Project No. 20231313

{
    \small
    \bibliographystyle{ieeenat_fullname}
    \bibliography{main}
}

\clearpage
\setcounter{page}{1}


\onecolumn
\begin{center}
{\Large \textbf{Appendix}}
    
    \vspace{\baselineskip} 
    
    
    
\end{center}

\section*{A. Quantitative comparison results for remaining tasks}

\begin{figure}[H]
    \centering
    \includegraphics[width=\columnwidth]{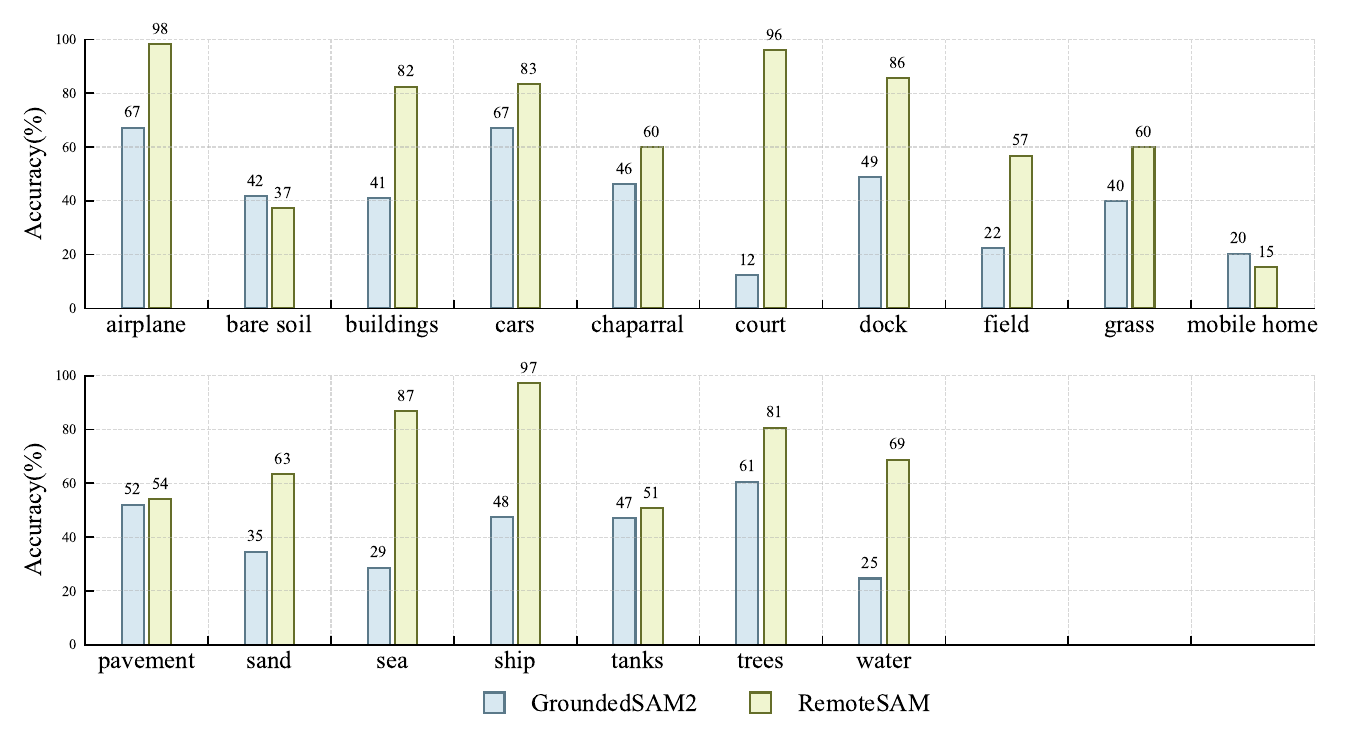}
    \caption{Detailed comparison with GroundedSAM2 on UCM\_Multilabel in SATIN Complex Scenes Task}
\label{fig:UCM}
\end{figure}

\begin{figure}[H]
    \centering
    \includegraphics[width=\columnwidth]{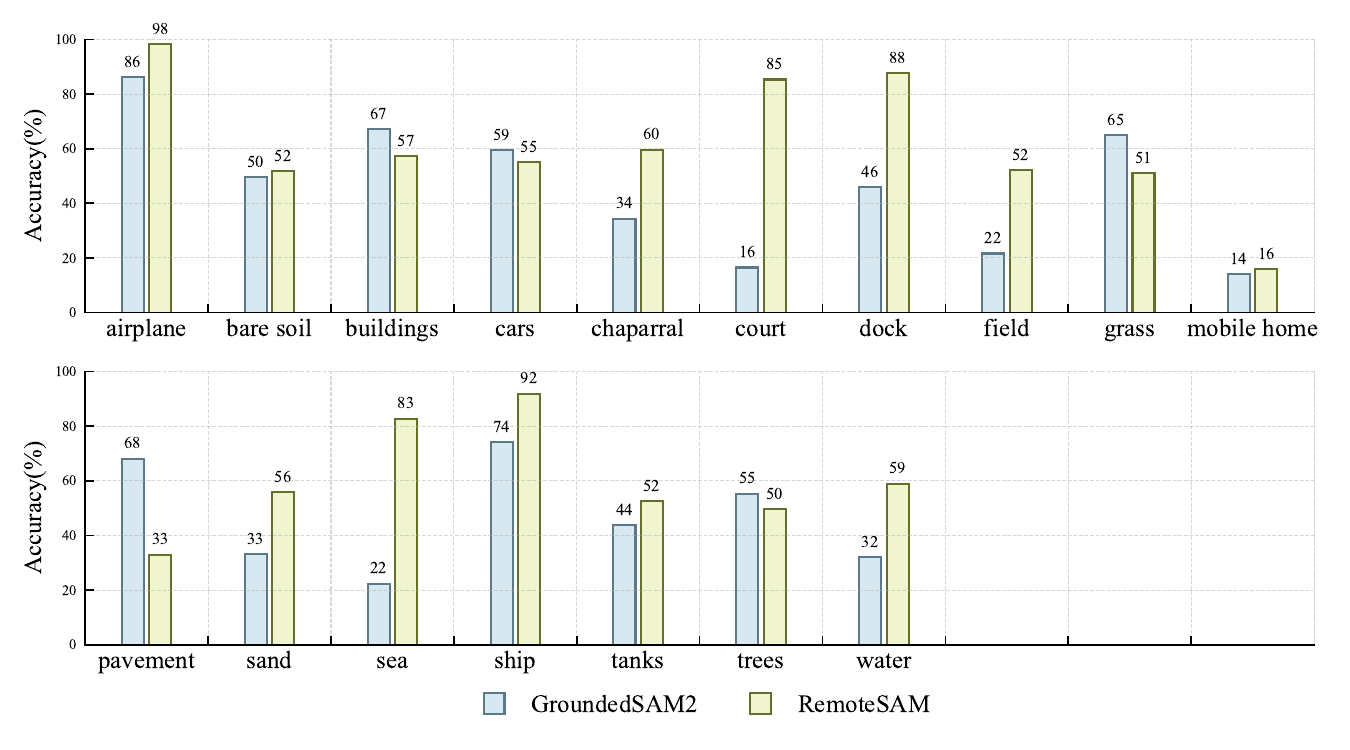}
    \caption{Detailed comparison with GroundedSAM2 on AID\_Multilabel in SATIN Complex Scenes Task}
\label{fig:AID}
\end{figure}

\begin{figure}[H]
    \centering
    \includegraphics[width=\columnwidth]{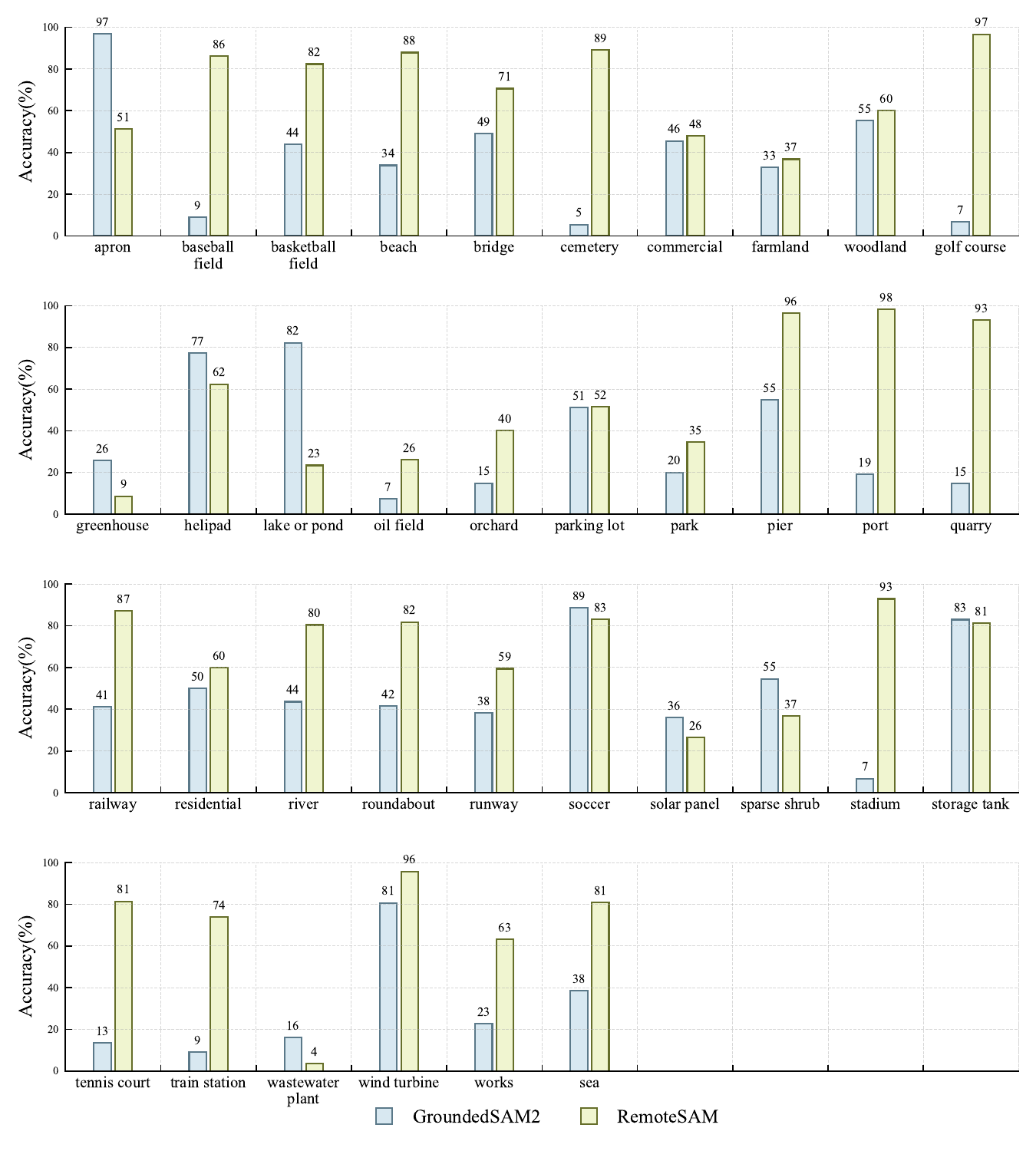}
    \caption{Detailed comparison with GroundedSAM2 on MultiScene in SATIN Complex Scenes Task}
\label{fig:MultiScene}
\end{figure}

\begin{figure}[H]
    \centering
    \includegraphics[width=\columnwidth]{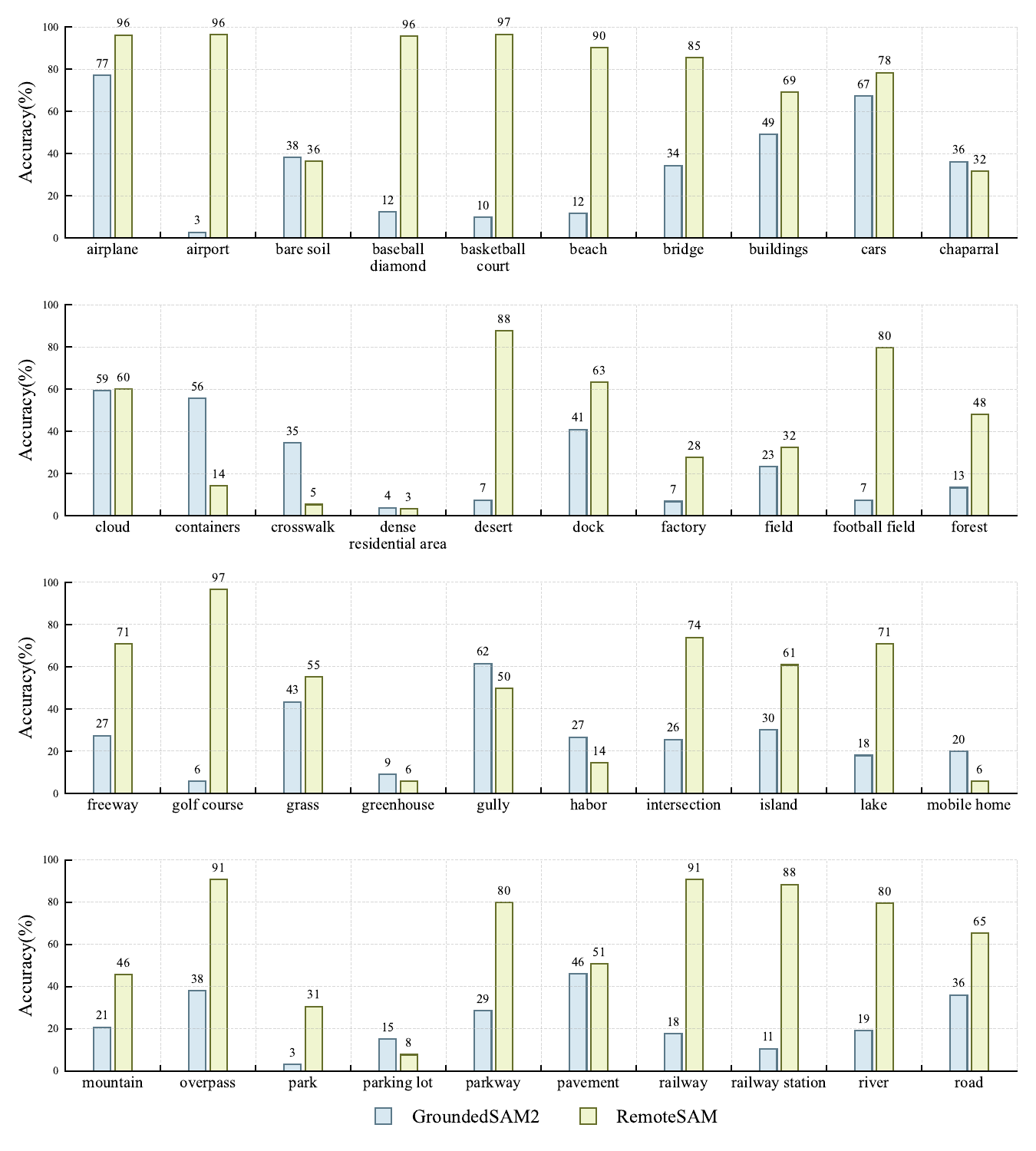}
    \caption{ Detailed comparison with GroundedSAM2 on MLRSNet in SATIN Complex Scenes Task (Part I)}
\label{fig:MLRSNetI}
\end{figure}

\begin{figure}[H]
    \centering
    \includegraphics[width=\columnwidth]{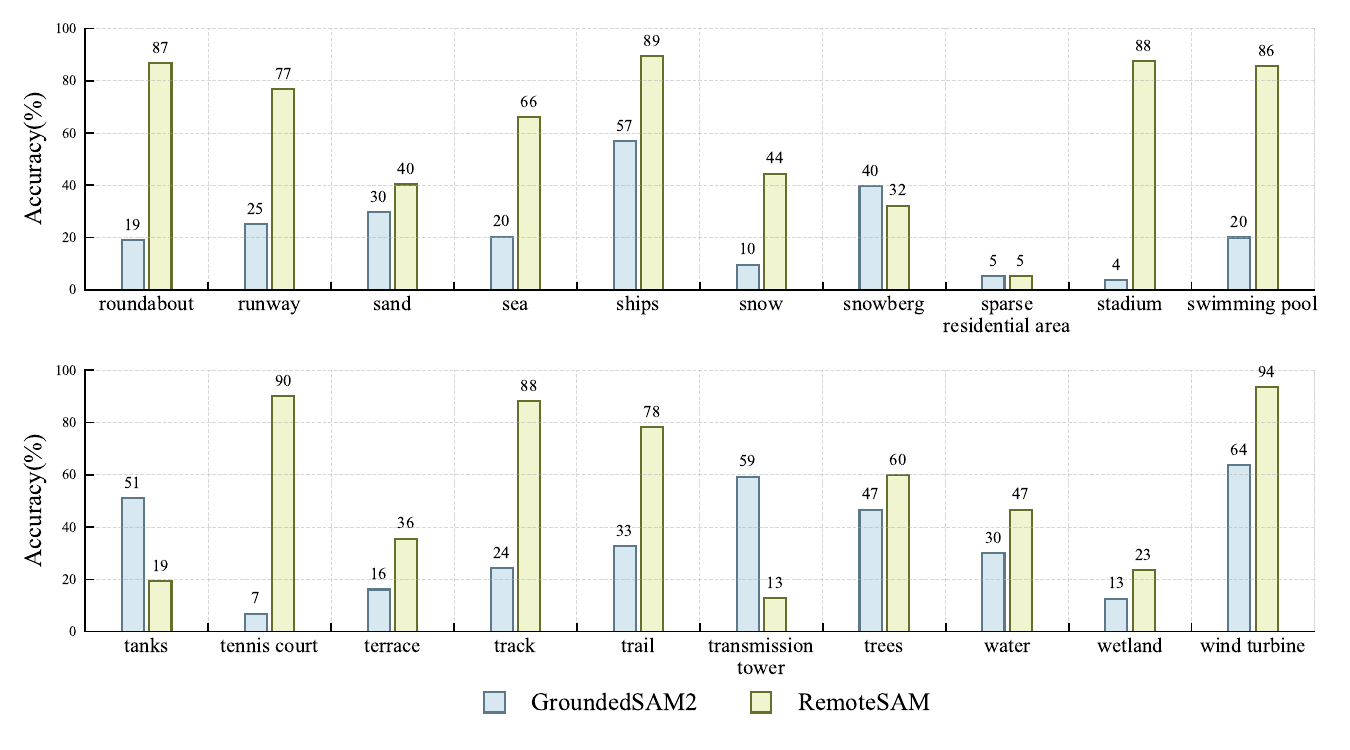}
    \caption{ Detailed comparison with GroundedSAM2 on MLRSNet in SATIN Complex Scenes Task (Part II)}
\label{fig:MLRSNetII}
\end{figure}

In this section, we present the performance of RemoteSAM for remaining tasks, including Referring Segmentation (c.f. Tab.~\ref{tab:RisBench}-Tab.~\ref{tab:RefSegRS}), Image Caption (c.f. Tab.~\ref{tab:CAP}) and Object Detection (c.f. Tab.~\ref{tab:HBB}-Tab.~\ref{tab:OBB}). Detailed comparisons of Complex Scenes Classification task in SATIN meta dataset are also depicted (c.f. Fig.~\ref{fig:UCM}-Fig.~\ref{fig:MLRSNetII}).

As shown in Tables ~\ref{tab:RisBench} to ~\ref{tab:RefSegRS}, RemoteSAM demonstrates exceptional performance on Referring Segmentation tasks, achieving top results across nearly all evaluated metrics on the three datasets, with only minor deviations in a few cases. For Image Caption task in Tab.~\ref{tab:CAP}, our rule-based captioning strategy attains a competitive CIDEr score of 12.370 on the UCM-Captions dataset compared to generic foundation models. Furthermore, RemoteSAM also excels in Object Detection tasks, achieving $AP_{50}$ scores of 62.74\% and 34.41\% on the DIOR and iSAID datasets, respectively, surpassing other foundation models. Besides, we also present category-level comparisons between RemoteSAM and GroundedSAM2, which strongly evidences that scaled semantic coverage does translate to improved generalization capability.

\begin{table}[H]
\centering
\caption{Detailed comparison of Referring Segmentation performance on RisBench.}
\setlength{\tabcolsep}{3mm}{
\begin{tabular}{l|c|ccccccc}
\toprule
\multirow{2}{*}{Methods} & \multirow{2}{*}{Publication} & \multicolumn{7}{c}{RisBench}                                                                               \\ \cline{3-9}

                         &                              & $Pr@0.5$ & $Pr@0.6$ & $Pr@0.7$ & $Pr@0.8$ & $Pr@0.9$ & $oIoU (\%)$ & $mIoU (\%)$  \\
\hline
BRINet                   & CVPR20                            & 52.87         & 45.39         & 38.64         & 30.79         & 11.86         & 48.73         & 42.91      \\
LSCM                    & ECCV20                        & 55.26         & 47.14         & 40.10         & 33.29         & 13.91         & 50.08         & 43.69      \\
CMPC                     & CVPR20                           & 55.17         & 47.84         & 40.28         & 32.87         & 14.55         & 50.24         & 43.82      \\
CMPC+                   & TPAMI21 & 58.02 & 49.00 & 42.53 & 35.26 & 17.88 & 53.98 & 46.73 \\
LAVT                     & CVPR22                        & 69.40          & 63.66         & 56.10          & 44.95         & 25.21         & 74.15         & 61.93      \\
CroosVLT        & TMM23 & 70.62 & 65.05 & 57.40 & 45.80 & 26.10 & 74.33 & 62.84 \\
CARIS          & MM23  & 73.94 & 68.93 & 62.08 & 50.31 & 29.08 & 75.10 & 65.79\\
LGCE                     & TGRS24                       & 69.64         & 64.07         & 56.26         & 44.92         & 25.74         & 73.87         & 62.13      \\
CroBIM                   & TGRS24                            & \underline{77.55}         & \underline{72.83}         & \underline{66.38}         & \underline{55.93}         & \underline{34.07}         & 73.04         & \underline{69.33}      \\
robust-ref-seg & TIP24  & 69.15 & 63.24 & 55.33 & 43.27 & 24.20 & 74.23 & 61.25 \\
RMSIN                     & CVPR24                           & 76.32         & 71.53         & 64.83         & 54.2          & 33.76         & \underline{75.24}         & 68.25      \\

\hline  \rowcolor{blue!10}
\textbf{RemoteSAM}            & -                         & \textbf{79.16 }& \textbf{74.24} & \textbf{67.74} & \textbf{58.09} & \textbf{38.80} & \textbf{75.93} & \textbf{71.46}      \\
\bottomrule
\end{tabular}
}
\label{tab:RisBench}
\end{table}

\begin{table}[H]
\centering
\caption{Detailed comparison of Referring Segmentation performance on RRSISD.}
\setlength{\tabcolsep}{3mm}{
\begin{tabular}{l|c|ccccccc}
\toprule
\multirow{2}{*}{Methods} & \multirow{2}{*}{Publication} & \multicolumn{7}{c}{RRSISD}                  \\ \cline{3-9}

                         &                              & $Pr@0.5$ & $Pr@0.6$ & $Pr@0.7$ & $Pr@0.8$ & $Pr@0.9$ & $oIoU (\%)$ & $mIoU (\%)$ \\
\hline
BRINet                  & CVPR20                       & 56.90          & 48.77         & 39.12         & 27.03         & 8.73          & 69.88         & 49.65         \\
LSCM                    & ECCV20                       & 56.02         & 46.25         & 37.7          & 25.28         & 8.27          & 69.05         & 49.92          \\
CMPC                     & CVPR20                       & 55.83         & 47.40          & 36.94         & 25.45         & 9.19          & 69.22         & 49.24    \\
CMPC+ & TPAMI21 & 57.95 & 48.31 & 37.61 & 24.33 & 7.94 & 70.13 & 50.12  \\
LAVT                     & CVPR22                       & 69.52         & 63.63         & 53.29         & 42.55         & 24.53         & 77.19         & 61.04      \\
CroosVLT        & TMM23 & 66.42 & 59.41 & 49.76 & 38.67 & 23.30 & 75.48 & 58.48 \\
CARIS           & MM23  & 71.50 & 63.52 & 52.92 & 40.94 & 23.90 & 77.17 & 62.12 \\
LGCE                      & TGRS24                       & 67.65         & 61.53         & 51.45         & 39.62         & 23.33         & 76.34         & 59.37      \\
CroBIM                    & TGRS24                         & 74.58         & 67.57         & 55.59         & 41.63         & 23.56         & 75.99         & 64.46        \\
robust-ref-seg  & TIP24 & 66.59 & 59.58 & 49.93 & 38.72 & 23.30 & 77.40 & 58.91 \\
RMSIN                    & CVPR24                       & 74.69         & 68.40          & 56.54         & 42.95         & 24.76         & 77.88         & 64.26       \\

RS2-SAM2                 & arXiv25                      & \underline{77.56}         & \underline{72.34}         & \underline{61.76}         & \underline{47.92}         & \textbf{29.73}         & \underline{78.99}         & \underline{66.72}          \\
\hline  \rowcolor{blue!10}
\textbf{RemoteSAM}            & -                          &\textbf{84.46} & \textbf{78.45} & \textbf{66.25} & \textbf{49.32} & \underline{29.62} & \textbf{80.04} & \textbf{71.75}      \\
\bottomrule
\end{tabular}
}
\label{tab:RRSISD}
\end{table}

\begin{table}[H]
    \centering
    \caption{Detailed comparison of Referring Segmentation performance on RefSegRS.}
    \setlength{\tabcolsep}{3mm}{
        \begin{tabular}{l|c|ccccccc}
        \toprule
        \multirow{2}{*}{Methods} & \multirow{2}{*}{Publication} & \multicolumn{7}{c}{RefSegRS}  \\ \cline{3-9}
          &                      & $Pr@0.5$ & $Pr@0.6$ & $Pr@0.7$ & $Pr@0.8$ & $Pr@0.9$ & $oIoU (\%)$ & $mIoU (\%)$ \\
        \hline  
        BRINet       &  CVPR20   & 22.56    & 15.74    & 9.85     & 3.52     & 0.5      & 60.16       & 32.87  \\
        LAVT         &  CVPR22   & 71.44    & 57.40     & 32.14    & 15.41    & 4.51     & \underline{76.46}       & 57.74  \\
        LGCE         &  TGRS24   & 73.75    & 61.14    & \underline{39.46}    & \underline{16.02}    & \textbf{5.45}     & \textbf{76.81}       & \underline{59.96}  \\
        CroBIM       &  TGRS24   & \underline{75.89}    & \underline{61.42}    & 34.07    & 12.99    & 2.75     & 72.33       & 59.77  \\
        RMSIN        &  CVPR24   & 68.63    & 52.61    & 26.47    & 10.13    & 1.82     & 71.46       & 55.71  \\
        \hline  \rowcolor{blue!10}
        \textbf{RemoteSAM}	 &  -        & \textbf{79.69}    & \textbf{70.89}    & \textbf{54.93}    & \textbf{24.11}    & \underline{5.01}     & 75.49       & \textbf{65.79}                   \\
        
        \bottomrule
        \end{tabular}
    }
\label{tab:RefSegRS}
\end{table}

\begin{table}[H]
    \centering
    \caption{Comparison of zero-shot Image Caption performance with Foundation Models.}
    \setlength{\tabcolsep}{3mm}{
        \begin{tabular}{l|c|c}
        \toprule
        \multirow{2}{*}{Methods} & \multirow{2}{*}{Publication} & UCM-Captions  \\ \cline{3-3}
                                &                              & CIDEr         \\
                                
        \hline  
        MiniCPM-V                & arXiv24                      & 0.000         \\
        MiniGPT-v2               & arXiv23                      & \textbf{16.282}        \\
        LLaVa-1.5                & NIPS24                       & 0.004         \\
        Qwen-VL-Chat             & arXiv23                      & 12.992        \\
        Florence-2-L             & CVPR24                       & \underline{13.844}        \\
        Sphinx                   & arXiv23                      & 0.000         \\
        Geochat                  & CVPR24                       & 0.288         \\
        LHRS-Bot                 & ECCV24                       & 8.365         \\

        \hline  \rowcolor{blue!10}
        \textbf{RemoteSAM}                     & -                            & 12.370             \\
        \bottomrule
        \end{tabular}
    }
\label{tab:CAP}
\end{table}

\begin{table}[H]
    \centering
    \caption{Comparison with Foundation Models of Object Detection with horizontal bounding box.}
    \setlength{\tabcolsep}{3mm}{
        \begin{tabular}{l|c|ccc}
        \toprule
        \multirow{2}{*}{Methods} & \multirow{2}{*}{Publication} & DIOR & iSAID & DOTAv2  \\ \cline{3-5}
                                &                              & \multicolumn{3}{c}{$AP_{50}(\%)$}     \\
        \hline  
        
        MiniGPT-v2	             & arXiv23                      & 9.430  & 2.54  & 1.62  \\
        Florence-2-L	         & CVPR24                       & 26.98 & 16.67 & 12.25 \\
        Qwen-VL-chat	         & arXiv23                      & 15.81 & 4.29  & 2.97  \\
        Sphinx	                 & arXiv23                      & 0.47  & 0.11  & 0.05  \\
        Falcon                   & arXiv25                      & \underline{56.65} & \underline{33.85} & \textbf{27.04} \\

        \hline  \rowcolor{blue!10}
        \textbf{RemoteSAM}       & -                            & \textbf{62.74}	 & \textbf{34.41} & \underline{20.17}       \\
        \bottomrule
        \end{tabular}
    }
\label{tab:HBB}
\end{table}

\begin{table}[H]
    \centering
    \caption{Object Detection performance with oriented bounding box of RemoteSAM.}
    \setlength{\tabcolsep}{3mm}{
        \begin{tabular}{l|c|ccc}
        \toprule
        \multirow{2}{*}{Methods} & \multirow{2}{*}{Publication} & DIOR  & iSAID & DOTAv2  \\ \cline{3-5}
                                &                              & \multicolumn{3}{c}{$AP_{50}(\%)$}     \\
        \hline  
        
        Falcon                   & arXiv25                    & \textbf{55.29} & \underline{28.83} & \textbf{23.29}  \\

        \hline  \rowcolor{blue!10}
        \textbf{RemoteSAM}     & -                            & \underline{55.22} & \textbf{34.58} & \underline{18.21}  \\
        \bottomrule
        \end{tabular}
    }
\label{tab:OBB}
\end{table}

\clearpage

\section*{B. Qualitative results}

In this section, we present the visualized results for each task as follows.

\subsection*{B.1. Task1: Referring Segmentation}

\begin{figure}[H]
    \centering
    \includegraphics[width=0.81\columnwidth]{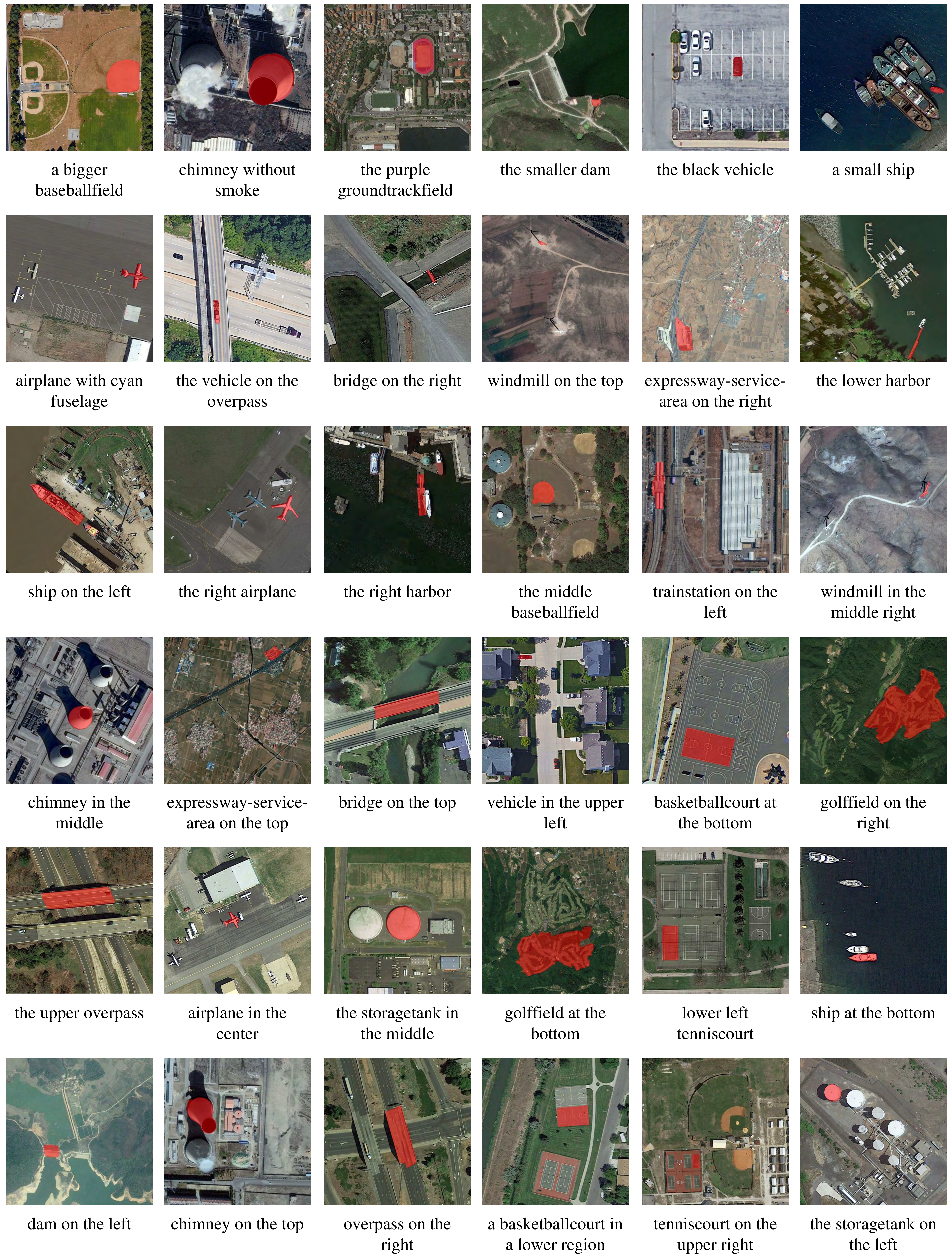}
    \caption{Task1: Referring Segmentation}
\label{Fig:REF}
\end{figure}

\clearpage

\subsection*{B.2. Task2: Semantic Segmentation}

\begin{figure}[H]
    \centering
    \includegraphics[width=0.99\columnwidth]{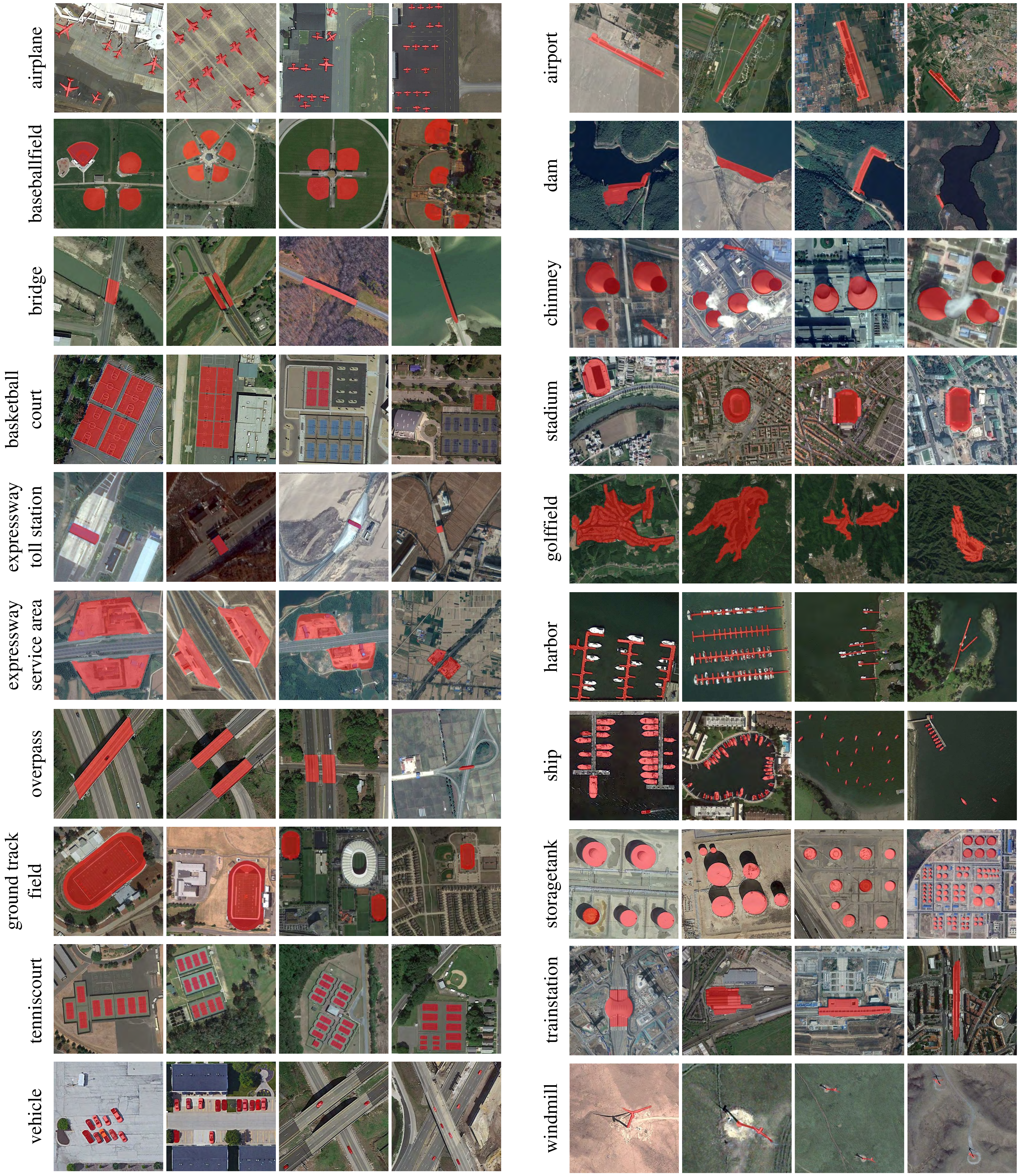}
    \caption{Task2: Semantic Segmentation}
\label{Fig:SEG}
\end{figure}

\clearpage

\subsection*{B.3. Task3: Object Detection}

\begin{figure}[H]
    \centering
    \includegraphics[width=0.98\columnwidth]{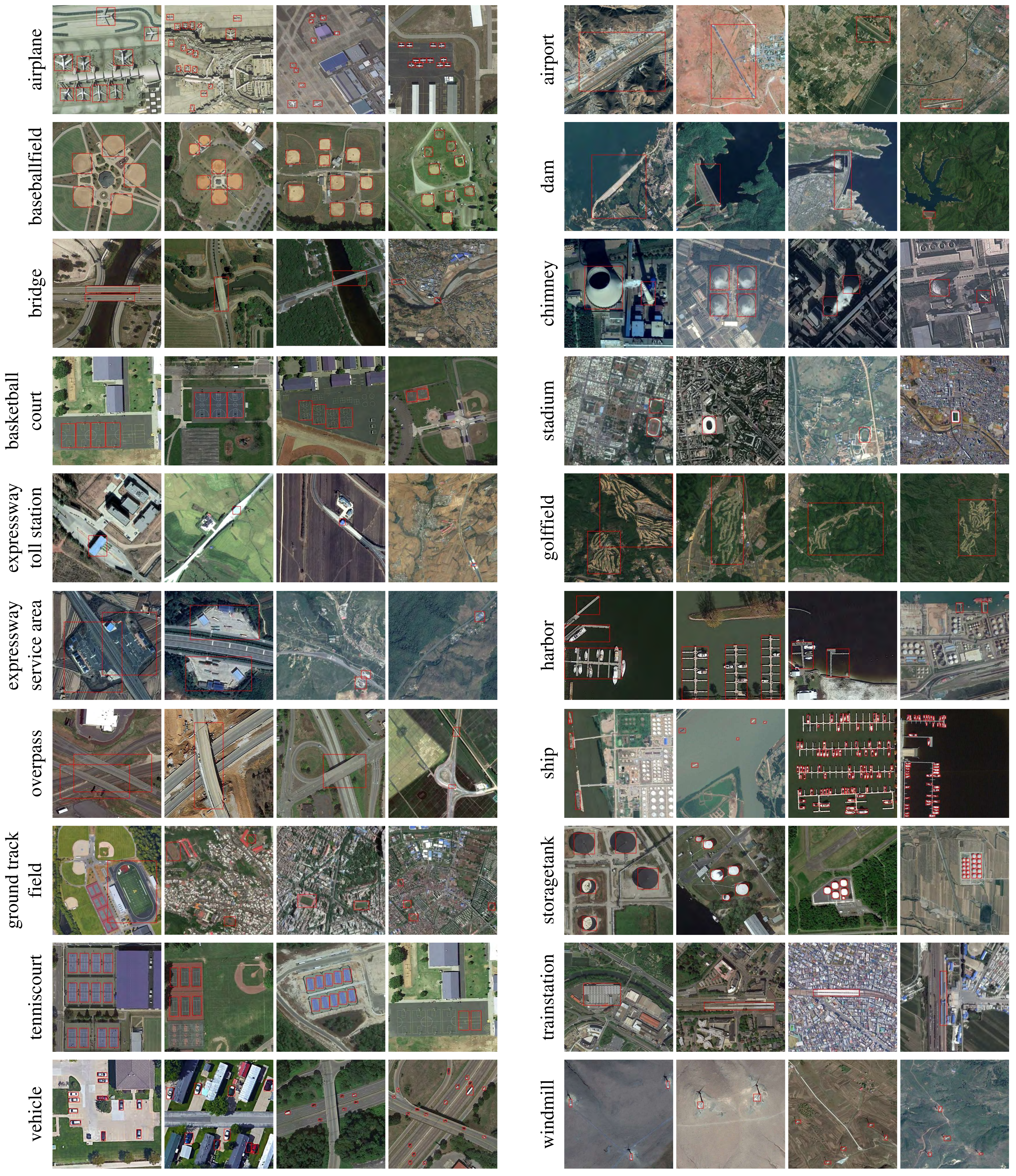}
    \caption{Task3: Object Detection}
\label{Fig:DET}
\end{figure}

\clearpage

\subsection*{B.4. Task4: Object Counting}

\begin{figure}[H]
    \centering
    \includegraphics[width=0.93\columnwidth]{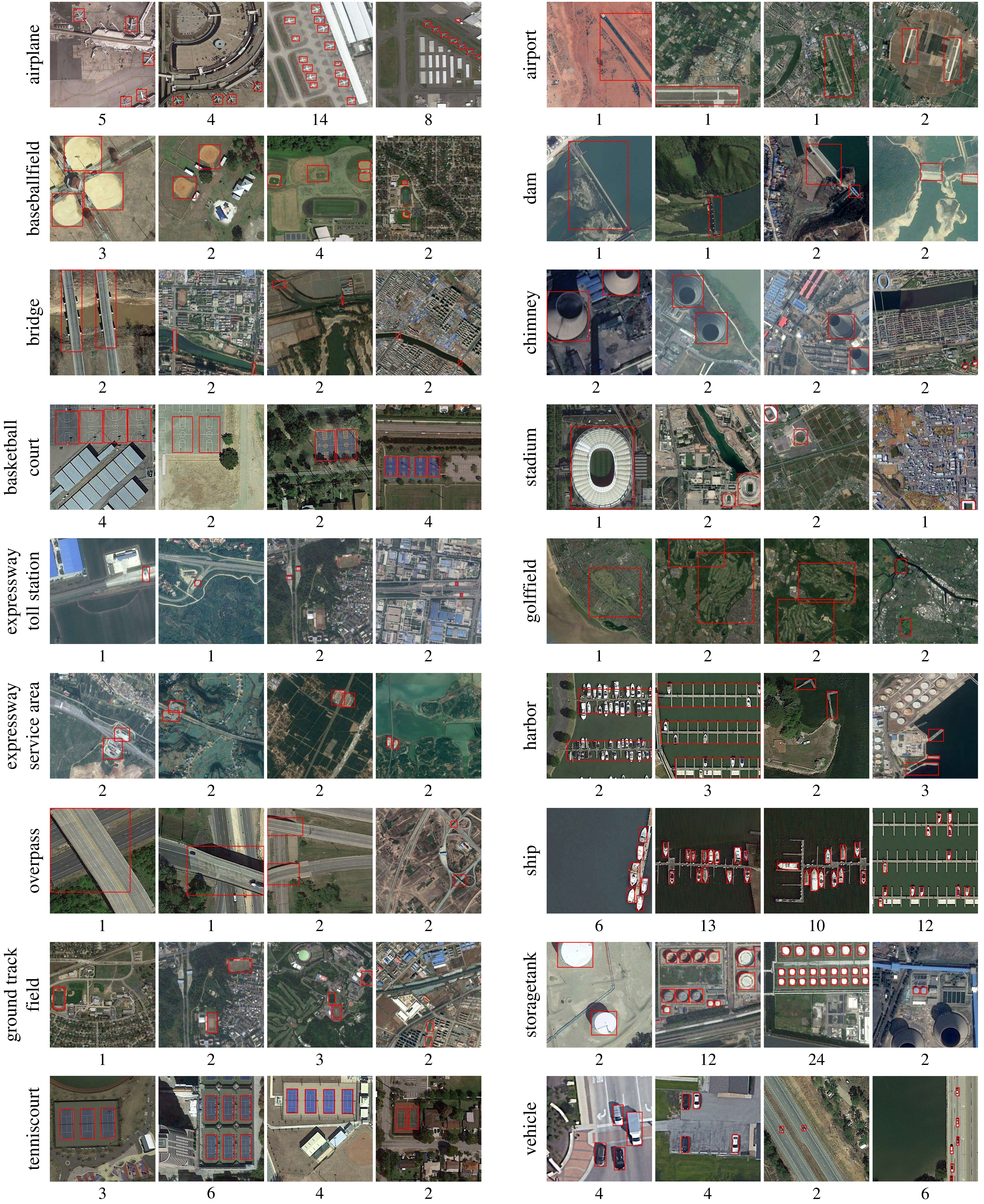}
    \caption{Task4: Object Counting}
\label{Fig:CNT}
\end{figure}

\clearpage

\subsection*{B.5. Task5: Visual Grounding}

\begin{figure}[H]
    \centering
    \includegraphics[width=0.86\columnwidth]{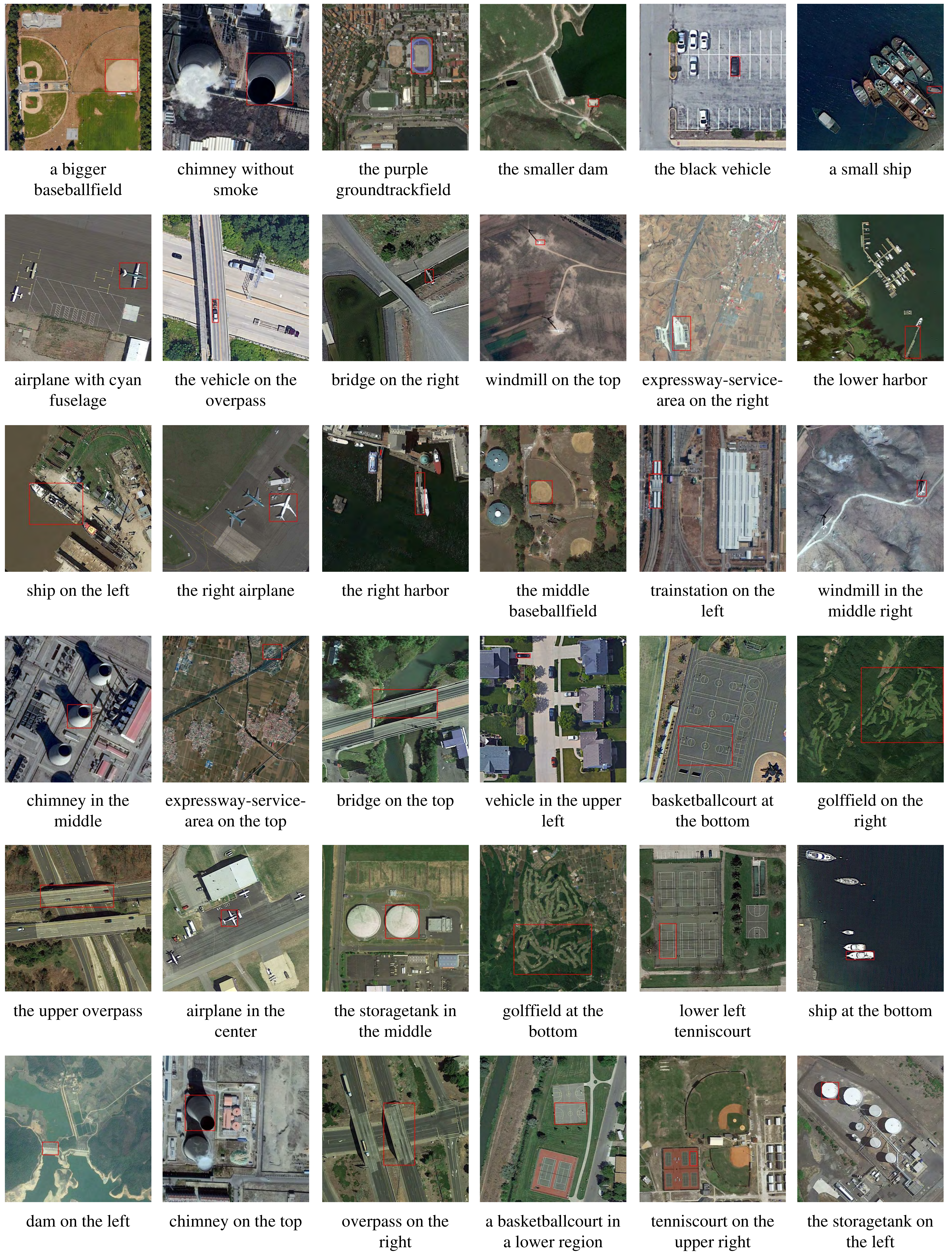}
    \caption{Task5: Visual Grounding}
\label{Fig:VG}
\end{figure}

\clearpage

\subsection*{B.6. Task6: Multi-Label Classification}

\begin{figure}[H]
    \centering
    \includegraphics[width=0.94\columnwidth]{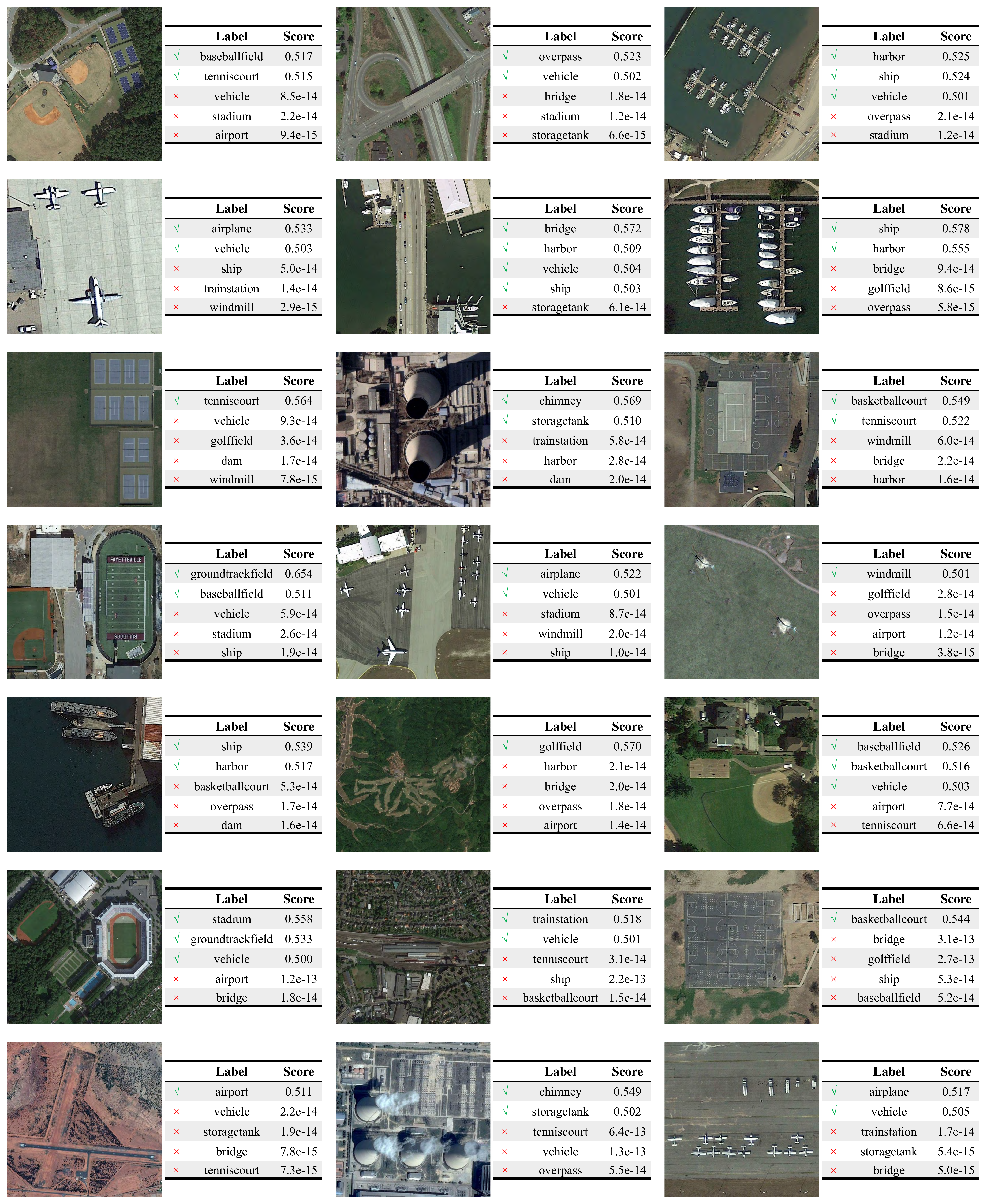}
    \caption{Task6: Multi-Label Classification}
\label{Fig:MLC}
\end{figure}

\clearpage

\subsection*{B.7. Task7: Image Classification}

\begin{figure}[H]
    \centering
    \includegraphics[width=0.95\columnwidth]{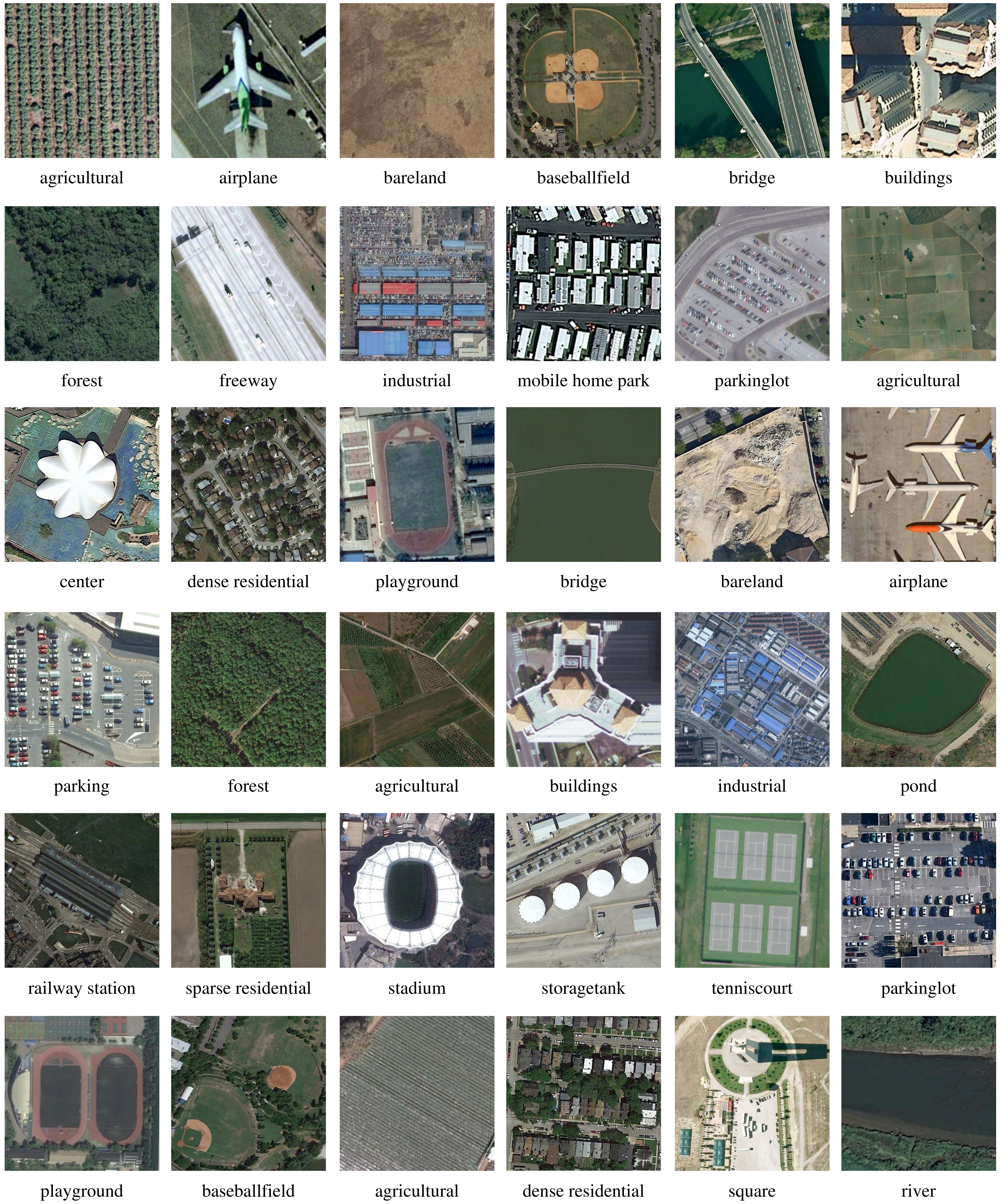}
    \caption{Task7: Image Classification}
\label{Fig:SC}
\end{figure}

\clearpage

\subsection*{B.8. Task8: Image Caption}

\begin{figure}[H]
    \centering
    \includegraphics[width=0.93\columnwidth]{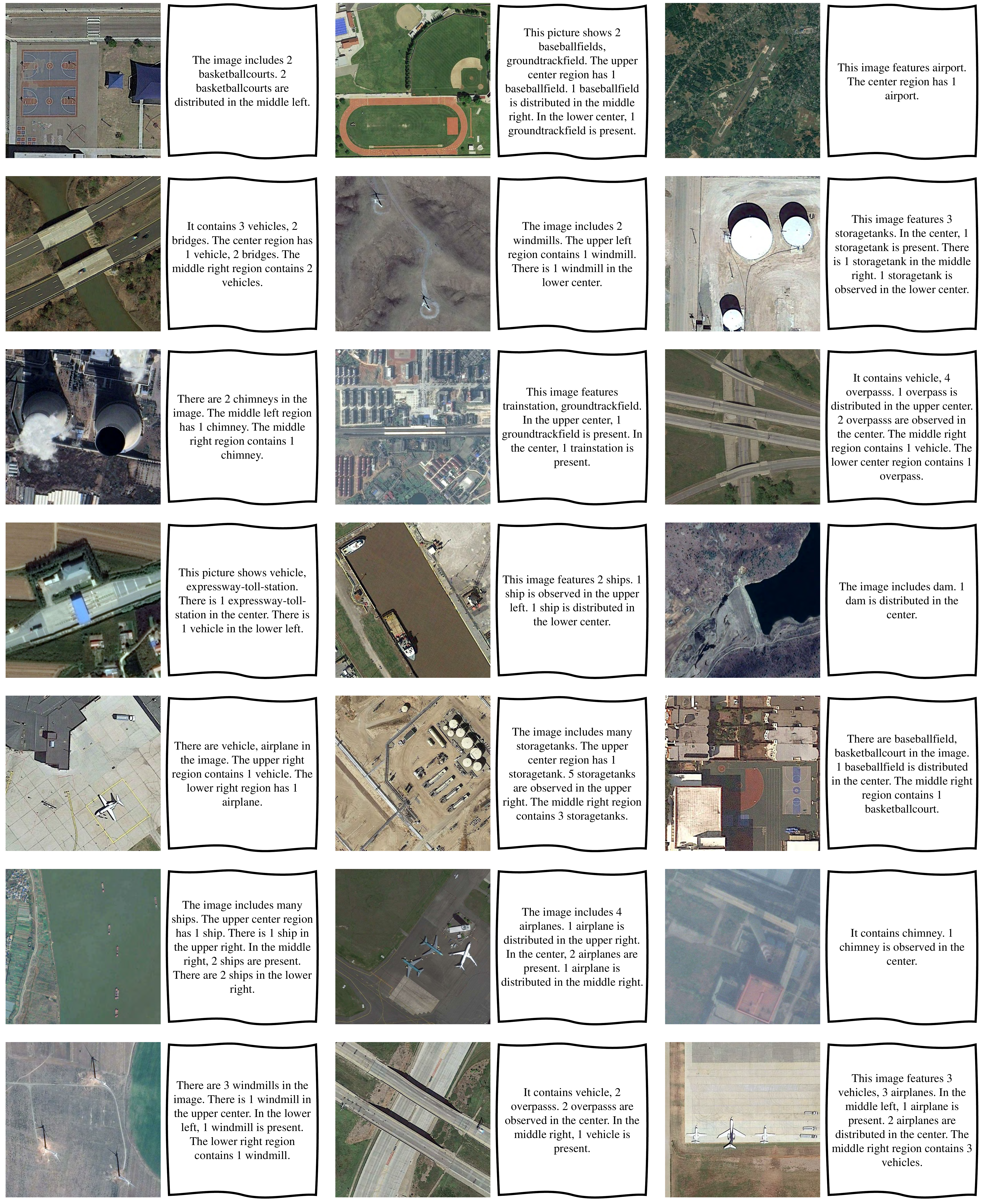}
    \caption{Task8: Image Caption}
\label{Fig:CAP}
\end{figure}

\clearpage

\section*{C. Ablation studies}

\textbf{Backbone Ablation:} In Tab.~\ref{ablation:backbone}, we investigate the effects of the text and image backbones through ablation experiments conducted on the RRSISD dataset with RemoteSAM. The combination of BERT as the text encoder and Swin-Base as the image encoder yields the highest performance, achieving an oIoU of 76.21\% and an mIoU of 64.79\%. 

In contrast, substituting the text encoder with Transformer while retaining Swin-Base results in a slight decrease in performance, with scores of 75.51\% for oIoU and 63.00\% for mIoU. This indicates that while Transformer remains competitive, BERT provides a marginally better contextual representation for Referring Expression Segmentation. Moreover, when ConvNext-B was employed as the image encoder, the performance further declined across both text encoders, suggesting that ConvNext-B may not capture the spatial hierarchies as effectively as Swin-Base in the context.

As a result, we select BERT as the text encoder and Swin-Base as the image encoder for our model, furnishing a powerful foundational computation unit for task unification.

\begin{table}[H]
    \centering
    \caption{Ablation study on Different Backbone}
    \setlength{\tabcolsep}{3mm}{
        \begin{tabular}{cc|cc}
        \toprule
        \multirow{2}{*}{Text Encoder} & \multirow{2}{*}{Image Encoder} & \multicolumn{2}{c}{RRSISD}  \\ \cline{3-4}
          &                      & $oIoU(\%)$ & $mIoU(\%)$ \\
        \hline  
        
        BERT            & Swin-B      & \textbf{76.21}  & \textbf{64.79}    \\
        Transformer     & Swin-B      & \underline{75.51}  & \underline{63.00}    \\
        BERT            & ConvNext-B  & 73.77  & 61.75    \\
        Transformer     & ConvNext-B  & 73.51  & 60.91    \\
        
        \bottomrule
        \end{tabular}
    }
\label{ablation:backbone}
\end{table}

\textbf{Effectiveness of CLIP Filtering:}  To validate the effectiveness of CLIP in filtering erroneous samples, we randomly sample 100 images from the filtered set and manually annotate their masks. As shown in the Tab.~\ref{ablation:quality}, the mIoU between the pseudo-labels and the manually annotated masks is 74.14\%. This result indicates that our reserved pseudo-labels are accurate.

\begin{table}[H]
    \centering
    \caption{Effectiveness of CLIP Filtering Strategy}
    \setlength{\tabcolsep}{3mm}{
        \begin{tabular}{l|ccccccc}
        \toprule
         Metrics & $Pr@0.5$ & $Pr@0.6$ & $Pr@0.7$ & $Pr@0.8$ & $Pr@0.9$ & $oIoU (\%)$ & $mIoU (\%)$  \\ 
        \hline  
         Pseudo-labels & 85.00 & 79.00 & 71.00 & 55.00 & 34.00 & 66.37 & 74.14 \\ 
        \bottomrule
        \end{tabular}
    }
\label{ablation:quality}
\end{table}

\textbf{Ablation of Multi-Label classification strategy :} In Tab.~\ref{ablation:clsstrategy}, we explore different strategies for Multi-Label classification. For mask-level strategy, a label is judged as positive when the area of its corresponding mask exceeds an area threshold(default as 0). While for the prob-level strategy, labels are considered positive when their confidence scores obtained from pooling surpass a threshold(default as 0.5). As shown in the Table, the prob-level strategy significantly outperforms the mask-level approach on both datasets.

\begin{table}[H]
    \centering
    \caption{Ablation of Multi-Label classification strategy}
    \setlength{\tabcolsep}{3mm}{
        \begin{tabular}{c|cc}
        \toprule
        \multirow{2}{*}{Strategy} & DIOR  & DOTAv2  \\ \cline{2-3}
                                  & $Acc(\%)$ & $Acc(\%)$ \\
        \hline  
        
        Mask-level                  & 92.708  & 66.774    \\
        Prob-level           & \textbf{94.042}  & \textbf{75.752}    \\
        
        \bottomrule
        \end{tabular}
    }
\label{ablation:clsstrategy}
\end{table}

\begin{figure}[H]
    \centering
    \includegraphics[width=0.95\columnwidth]{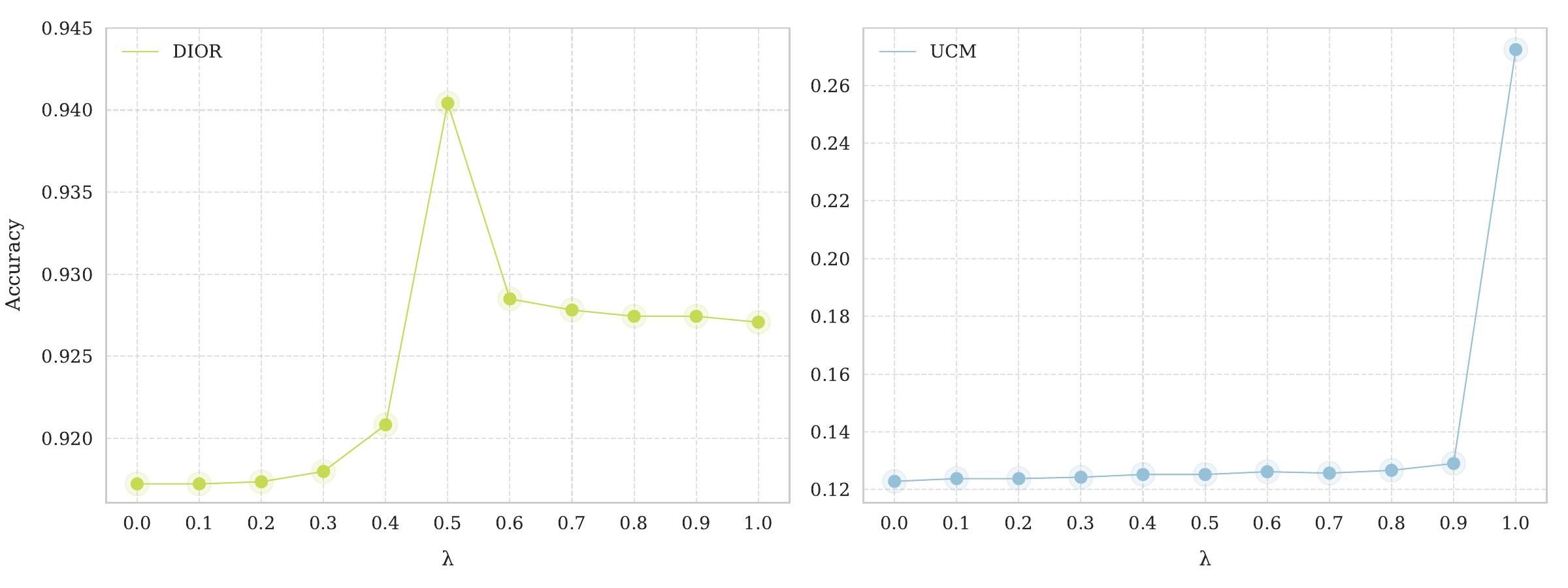}
    \caption{Balance factor analysis of classification}
\label{ablation:hyperpara}
\end{figure}

\textbf{Balance factor analysis of classification:} To ascertain the optimal balance factor values for classification, we conduct experiments by varying the balance factor $\lambda$ within the lass-wise probability aggregation function (Eq.3). The experimental results are illustrated in Fig.~\ref{ablation:hyperpara}. The results indicate that when $\lambda$ is set to 0.5 and 1, Multi-Label Classification and Image Classification achieved the highest accuracy, respectively. Therefore, we adopt this set of parameters for our final test results.

\begin{figure}[H]
    \centering
    \includegraphics[width=0.95\columnwidth]{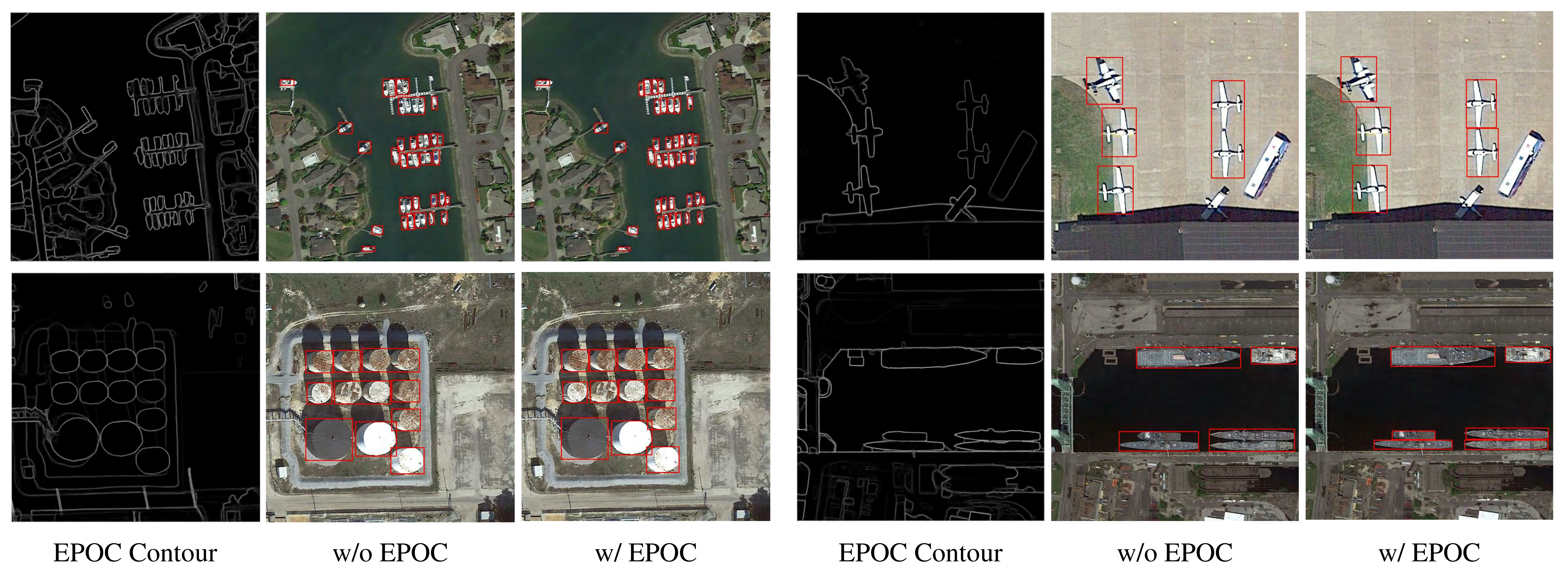}
    \caption{Qualitative examples of EPOC refinement in Object Detection}
\label{ablation:epoc}
\end{figure}

\textbf{Qualitative examples of EPOC refinement in Object Detection: } As illustrated in Fig.~\ref{ablation:epoc}, the refinement by EPOC effectively resolves the limitations of the M2B strategy in processing adjacent targets. 


\clearpage
\section*{D. Examples of RemoteSAM-270k}
\begin{figure}[H]
    \centering
    \includegraphics[width=0.93\columnwidth]{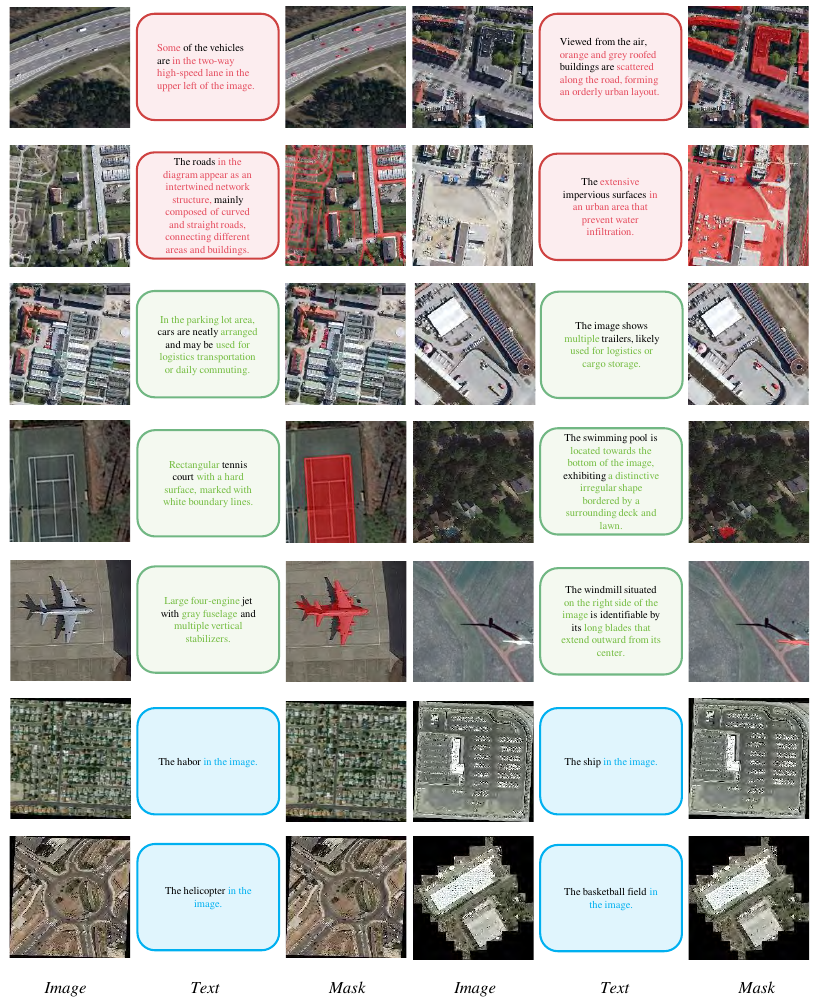}
    \caption{Examples and categories of RemoteSAM-270k}
\label{Fig:270k}
\end{figure}

\section*{E. Prompts Setting of Expressions Creation for Qwen2.5VL}
\begin{table}[H]
    \centering
    \caption{An Example implementation of extracting class-text pairs by prompting Qwen2.5VL.}
    \begin{tcolorbox}[title = {Prompt for Extracting Class-Text Pairs},
        fonttitle = \bfseries, fontupper = \sffamily, fontlower = \itshape]
        \hspace*{-0.25cm}
        \begin{tabular}{>{\raggedright\arraybackslash}m{1\textwidth}}
            \textcolor{blue!50!black}{\textbf{System Message:}} \\
            \addlinespace[3pt]
            \textcolor{black}{\textbf{Input:}} 
            \begin{itemize}
                \item You will receive a detailed image caption of a remote sensing image.
            \end{itemize} 
            \textcolor{black}{\textbf{Task Objective:}}
            \begin{itemize}
                \item Please extract all Object categories from the caption.
                \item Please generate a new concise description for each category.
            \end{itemize} 
            \textcolor{black}{\textbf{Guidelines:}}
            \begin{itemize}
                \item Please focus on describing this single-class object and its attributes and spatial relation-ships.
                \item The caption must be brief, and no more than 20 words.
                \item The final output is in JSON format, with these categories as keys and corresponding descriptions as values.
            \end{itemize} \\
            \hline \\
            \textcolor{blue!50!black}{\textbf{User:}} \\
            \addlinespace[3pt]
            \textbf{Detailed Caption:} This image is an aerial view of an airport terminal and its surroundings. There are numerous gates around the terminal, each connected to the main building via jet bridges. The apron area is extensive, with many aircraft parked at the gates...\\
            \addlinespace[6pt]
            \hline \\
            \textcolor{blue!50!black}{\textbf{Assistant:}}\\
            \addlinespace[3pt]
            \textcolor{black}{\textbf{Json Format Output:}}\\
            \{\\
            \hspace*{16pt}``aircraft'': ``Multiple aircraft parked on the apron.''\\
            \hspace*{16pt}``jet bridges'': ``Multiple jet bridges connecting gates to the terminal.''\\
            \hspace*{16pt}...\\
            \}
        \end{tabular}
    \end{tcolorbox}
\label{prompt}
\end{table}

\end{document}